%% file: main.tex
\pdfoutput=1

\documentclass{article}

\usepackage{arxiv}

\usepackage[utf8]{inputenc} 
\usepackage[T1]{fontenc}    
\usepackage{hyperref}       
\usepackage{url}            
\usepackage{booktabs}       
\usepackage{amsfonts}       
\usepackage{nicefrac}       
\usepackage{microtype}      
\usepackage{lipsum}		
\usepackage{graphicx}

\usepackage{subfig}  

\usepackage{amsmath} 

\usepackage{booktabs} 
\usepackage{etoolbox}
\usepackage{xspace}
\usepackage{datetime}
\usepackage{tabularx}
\usepackage{fancyvrb}
\usepackage{enumitem}
\usepackage[ruled,vlined,linesnumbered]{algorithm2e}
\usepackage{letltxmacro}

\usepackage{pseudo} 

\pseudoset{  
st-left=‘, st-right=’, stfont=\textit,
ct-left=\texttt{/\!/}\,, ct-right=, ctfont=
}

\input{commands}

\title{Expedition: A System for the Unsupervised Learning of a
  Hierarchy of Concepts}

\author{Omid Madani\\ 
  Cisco Secure Workload \\
  456 University Ave., Suite 200 \\
  Palo Alto, CA 94301 \\
  omadani@cisco.com
}

\renewcommand{\headeright}{}
\renewcommand{\undertitle}{}

\hypersetup{
  pdftitle={Expedition: A System for the Unsupervised Learning of a
    Hierarchy of Concepts},
 pdfauthor={Omid Madani},
 pdfkeywords={Unsupervised learning, Learning concepts, Self-supervised learning}
}

\begin{document}
\maketitle

\input{abstract} 	

\vspace*{.5in}

\hspace*{0.7in}  \mbox{"Concepts are the glue that hold our mental world
together.", G. Murphy, The Big Book of Concepts.}


\hspace*{0.7in} \mbox{"..  to cut up each kind according to its species along
its natural joints, ...", Plato, Phaedrus.}

\input{intro} 	%
\input{overview} 
\input{concepts} 	%
\input{seg} 	%
\input{learning} 	%
\input{exps} 	%
\input{rel}

\input{summary}

\input{acks}



\bibliographystyle{plain}  

\bibliography{global}

\appendix

\input{appendix_rate}

\input{appendix_binary}

\end{document}

%% file: commands.tex
\newcommand{\msq}[1]{[#1]}

\newcommand{\con}{con} 

\newcommand{\coma}{COMA } 
\newcommand{\comap}{COMA} 

\newcommand{\eps}{\epsilon} %

\newcommand{\prior}{\mbox{prior}}

\newcommand{\score}{\mbox{score}}

\newcommand{\pred}{\mbox{prob}}

\newcommand{\mysim}{\raise.17ex\hbox{$\scriptstyle\sim$}}
\newcommand{\mytilde}{\raise.17ex\hbox{$\scriptstyle\sim$}}

\newcommand{\ie}{{\em i.e.~}} 
\newcommand{\vs}{{\em vs.~}}  
\newcommand{\eg}{{\em e.g.~}}

\newcommand{\co}[1]{}

\newcommand{\match}{\mbox{MatchReward}}  



%% file: abstract.tex
\begin{abstract}
We present a system for bottom-up cumulative learning of myriad
concepts corresponding to meaningful character strings (n-grams), and
their part-related and prediction edges.  The learning is
self-supervised in that the concepts discovered are used as predictors
as well as targets of prediction.
We devise an objective for segmenting with the learned concepts,
derived from comparing
to a baseline (reference) prediction system, that promotes making and using larger concepts,
which in turn allows for
predicting larger spans of text, and we describe a simple technique to
promote exploration, \ie trying out newly generated concepts in the
segmentation process.  We motivate and explain a layering of the
concepts, to help separate the (conditional) distributions learnt
among concepts.  The layering of the concepts roughly corresponds to a
part-whole concept hierarchy in this work.  With rudimentary segmentation and learning
algorithms, the system is promising in that it acquires many concepts
(tens of thousands in our small-scale experiments), and it learns to
segment text well: when fed with English text with spaces removed,
starting at the character level, much of what is learned respects word
or phrase boundaries, and over time the average number of "bad" splits
within segmentations, \ie splits inside words, decreases as larger
concepts are discovered and the system learns when to use them during
segmenting. We also report on promising experiments when the input
text is converted to binary and the system begins with only two concepts,
"0" and "1".  The system is transparent, in the sense that it is easy
to tell what the concepts learned correspond to, and which ones are
active in a segmentation, or how the system "sees" its input in an
episode.  We expect this framework to be extensible and we discuss the
current limitations and a number of directions for enhancing the
learning and inference capabilities.
\end{abstract}

%% file: intro.tex
\section{Introduction}

Concepts, such as "water", "chair", and "eat", are fundamental to
human intelligence: our mental model(s) of the world and our basic
cognition are founded on concepts \cite{bigbook}.  What is the nature of
concepts, \ie how are they computationally represented, and how can a
system acquire diverse and richly inter-related concepts in an
unsupervised manner, \ie without an explicit teacher, from the
low-level perceptual stream?  There is evidence that much learning, of
numerous concepts, and how they relate and constrain one another, and
ultimately a {\em sense} of what is probable or common in every day
experience, in humans and animals, takes place without explicit
teaching, achieved largely through observing
\cite{Sheridan73,law2011infant,catDev03,theories1}.  Inspired by considerations
of early human learning,
and in
particular perceptual and category learning, that likely continue
throughout life
\cite{kellman2009perceptual,Carvalho2016HumanPL,theories1}, here we
propose and explore an approach to efficient unsupervised learning of
{\em discrete and discernible} "concepts", in a sparse manner in the
text domain. The learning is achieved in a cumulative bottom up
fashion.

\begin{figure}[t]
\begin{center}
  \centering
  \hspace*{-.4in} \subfloat[System architecture. ]
          {{\includegraphics[height=6cm,width=12cm]{
        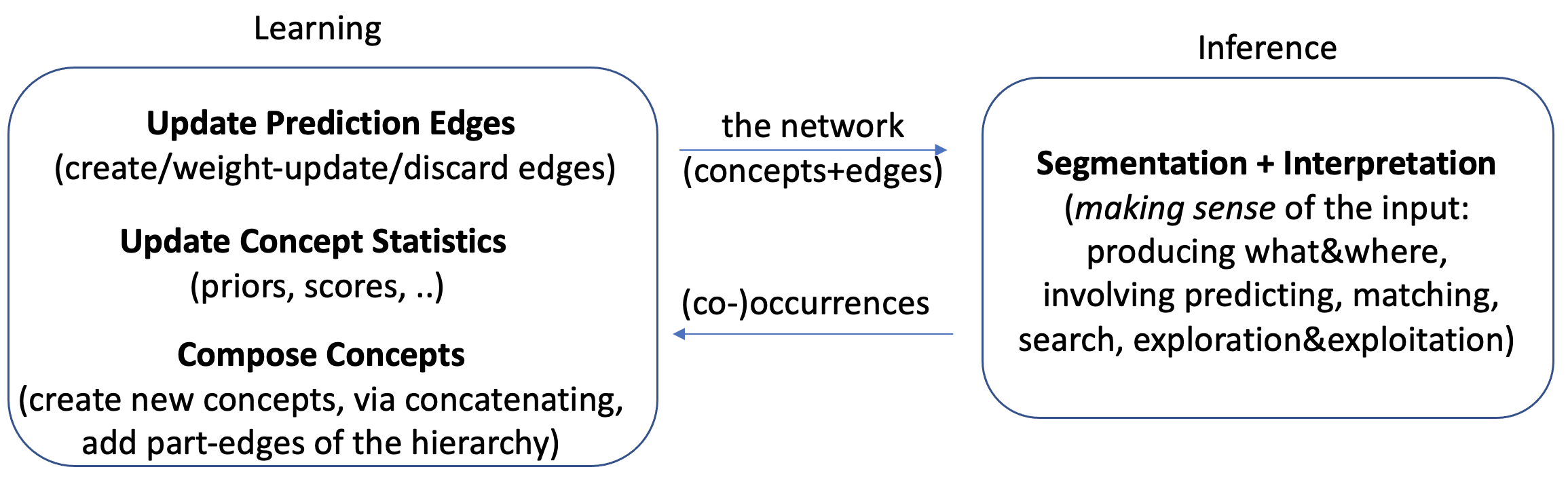} }}
  \hspace*{-.1in} \subfloat[The main learning loop.  ]
          {{\includegraphics[height=6cm,width=7cm]
              {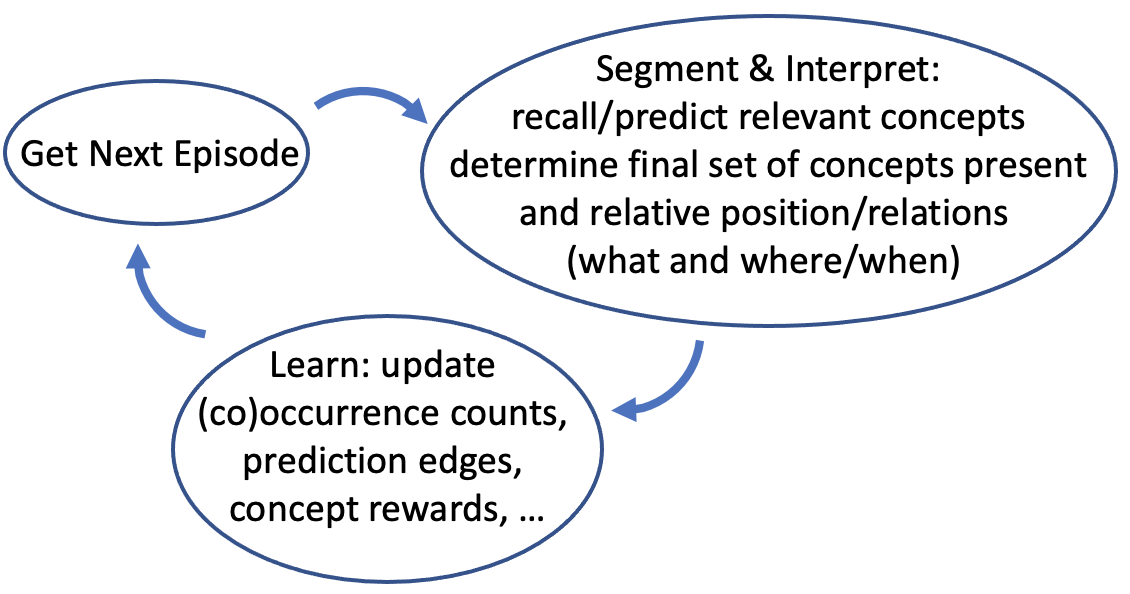}
           }}%
\end{center}
\caption{(a) The architecture of the Expedition system: Learning
  provides the (hierarchical) network of concepts to inference, and
  inference determines, in each episode, which (small) part of this
  network needs updating. (b) Repeatedly get an episode (a span of
  text), make "sense" of it (\ie summon the candidate relevant
  concepts, via recall/prediction, matching and coherence), then learn
  from the final selected construction (the final segmentation
  structure): using the final segmentation, update (co-)occurrence
  counts, prediction weights, etc.  The Expedition prediction system
  is a realization of this general learning loop.}
\label{fig:combo} %
\end{figure}

One way to motivate this particular work is as follows: Imagine a
system endowed with the capabilities of a human brain, but that has
access only to a stream of text, such as all the English language text
on the web. The system "lives in text" or has a single modality only,
\ie its percepts or senses are limited to detecting characters and
their relative positions in a limited span, such as up to 100
consecutive characters at a time.  Natural language is rich and
contains numerous types of regularities. The web is vast.  How much
can the system (sufficiently inclined) learn from this
information-rich and practically unlimited stream, and what learning
capabilities are activated? Will the system pick up words and phrases?
Will it learn, in effect, the rules of grammar and the structural cues
that can be derived from the punctuation, and will it pick up certain
patterns such as "A" and "a" are often interchangeable, and so are the
numerical digits?  How would it organize and structure the acquired
knowledge? Would internal discernible concepts develop and if so can
we establish a correspondence to the input text, such as words and
phrases, and perhaps general abstract categories such as the category
of verbs?  Would multiple such systems trained on sufficiently similar
input streams, share enough common concepts to communicate, and in
effect compare and adjust their concepts in a useful manner?  What
would the concept structure(s) and interconnections look like?  What
kind of changes take place in a single episode, and what guarantees
can we make regarding the evolution of long-term learning, as these
changes accumulate?  Can we build components of such learning
mechanisms in an artificial learning system?  Our work explores a
scalable unsupervised approach to learning simple concepts, involving
one inference and two main learning components, driven by the goal of
improving prediction in the text stream.

A concept in this work corresponds to a linear pattern, or a string of
characters, such as "a", "is", and "school".  However, a concept is
more than just the string pattern it corresponds to: it has
associations to other concepts, and may have parts and may be parts of
many other concepts. Thus a concept is a node in a network, and the
network contains different types of edges. In particular, the network
is hierarchical. We achieve learning of concepts via a system, we call
the {\bf \em Expedition} system, that contains several
(learning/inference) modules.  Figure \ref{fig:combo} gives a picture
of the different modules in the system and their interaction, as well
as the {\bf \em basic learning loop}.

Each concept maintains and updates statistics such as occurrence and
co-occurrence counts, and prediction weights, derived from the stream
of text that it processes.\footnote{It is natural to describe the
  various computations and datastructures, from the point of view of
  concepts, specially for the learning.  } The system uses the
predictions to {\bf \em segment} and {\bf \em interpret} its
input. This process breaks a span of text (a line, or a string of
characters) into chunks or subsegments, by simultaneously mapping the
subsegments into its existing concepts. Segmentation+interpretation
can be thought of as a rudimentary "sense making". We drop the term
"interpretation" and refer to this process as segmentation for short.
The segmentation of the string of characters and the mapping of the
segments into a few internal concepts, which we refer to as the {\bf
  \em active concepts}, provides the 'what' and 'where' of concepts in
an episode. This information is in turn used to update various
statistics such as the active concepts' prediction weights (predicting
one another) and co-occurrences. This main learning loop of
segmentation (sense-making) and updating weights repeats ad infinitum,
and is a major source of learning.

Another important form of learning is discovering new concepts. In
this work, new concepts are built via concatenating ({\bf \em
  composing}) existing concepts based on a few co-occurrence criteria,
where the co-occurrence information is collected over the many
learning episodes of the main learning loop.  As new concepts are
discovered and used during segmentation, prediction relations among
the concepts are added or updated, and the network grows.  In our
relatively small-scale experiments involving a few million episodes,
tens of thousands of concepts, corresponding to n-grams of characters
(words and phrases) are discovered.  Connections among concepts are
kept relatively sparse: a concept has relatively few edges compared to
the totality of concepts. In this work, the edges in the network are
{\bf \em prediction edges} and {\bf \em part-related
  edges}. Constraints from matching the raw input combined with how
well the active concepts predict one another, which we refer to as
{\bf \em \comap} (COherence and MAtch), help determine the final chosen segmentation in
an episode, and the chosen segmentation determine which concepts and
co-occurrences are observed and updated.  The concept co-occurrence
counts are used, over time, to compose bigrams of existing concepts
(an instance of the AND or the conjunction operation), and thus to
build new "larger" concepts (\ie concepts that correspond to larger
patterns in the input).  The larger concepts enable the system to be
more successful at predicting, in particular for successfully
predicting increasingly longer stretches of the input stream.




In summary, the system engages in a self-learning or a so-called {\em
  self-supervised} learning regime, setting up for itself many
learning episodes or puzzle solving sessions, to build a large
predictive model of its world.  Can higher-level concepts, \eg in our
case terms and phrases in natural language, be discovered effectively
in a bottom up manner, starting from the level of characters, via
repeated segmentation and prediction as we described?  There are many
challenges. There are several sources of uncertainty, imperfections,
and noise, including: 1) limits of the local window used for
prediction, 2) finite experience with concepts and tail challenges: at
any given point, many concepts are relatively new and may not have
reliable statistics or provide reliable predictions, 3) imperfections
of the segmentation process: the search algorithms make local moves
and can search only a tiny fraction of segmentation
possibilities. Thus, the proposed system is composed of several
interacting noisy modules.  A major question is therefore whether the
combination of prediction, composition (bigram creation in this work)
and segmentation effectively yields sufficiently good higher-level
concepts, since all the modules provide error-prone inputs to one
another. Any errors made may persist (\eg bad compositions that
continue to be used) or have the potential to compound and propagate
over time and may slow or stop progress (\ie convergence to and
getting stuck in local optima).

Another challenge is devising an appropriate objective to "motivate"
the system to build (discover) larger concepts. We propose an
objective based in part on roughly how large the concept is, and in
part on how well it is predicted (both historically on average, as
well as in a particular episode).  The objective is derived from
measuring how well the system is performing compared to simple
prediction at the character level.  Since concepts are discovered
incrementally, and we define their utility in terms of how well they
are predicted, we will also have to address the need for an
exploratory period: a newly generated concept needs to be used a few
times before we have an adequate estimate of its goodness as well as
when to use it.

The system, with rudimentary algorithms that we describe, is
promising: When run on natural language English text with blank
spaces removed, starting at the character level, we find that, after
some learning, the n-grams
learned, \ie the higher level
concepts, correspond to words and phrases,
and the splitting of the text into concepts continues to improve with
more training.  Thus we provide evidence that the combination of
prediction, composition, and segmentation can be an effective
self-supervised learning strategy.  In the end, the system is a kin to
a language model: it provides both a hierarchical vocabulary (of
concepts), as well as a way of segmenting the low level characters and
mapping them into this vocabulary. It does not require any
preprocessing of the text.  The only requirement is providing ample
(unlabeled) data.

Modern large language models based on deep neural networks, trained
using backpropagation, incorporate a number of advances and have found
a diverse range of applications
\cite{Brown2020LanguageMA,dong2019unified}. A major question is
whether and in what ways they can be extended to provide us the
(possibly discrete) concepts needed for advanced cognitive tasks.  On
the other hand, for our approach, to be competitive with such, a
number of enhancements are needed. Throughout the paper, we discuss
various challenges, possible alternatives to our current
implementation, and potential enhancements.  A major open question is
how much the learning representation power of this approach can be
extended, in particular to achieve some manner of abstraction (and how
this may be combined with composition), to, for example, further
address sparsity and tail challenges.  We discuss a few ideas in those
directions in the paper.  We expect that the approach is flexible and
the system is extensible, in particular for incorporating additional
kinds of learning.

This paper is a snapshot of our project for advancing research on
systems that acquire myriad concepts with rich relations in an
unsupervised manner. The models we report on are also snapshots in
another sense: the learning can keep going, but timing and computation
(memory) constraints have limited what we report on.  We provide
evidence that the learning continues to make good progress, and
highlight areas for improving the approach. We only briefly explore
some algorithmic and parameter variations (such as choice of learning
rate schedule) and interactions among parts, and we leave more
extensive comparisons and further explorations of the space to future
work.

We begin the paper by an overview of the Expedition system. We then
describe concepts and edges (the network structure) in Sect.
\ref{sec:cons}. We present the segmentation and prediction processes,
and scoring and objectives in Sect. \ref{sec:segs}, followed by
Sect. \ref{sec:learning} on learning (updating edges and learning new
concepts). We then present experiments, which includes various plots
of trajectories of learning, examples of what is learned and example
segmentations, together with some exploration of the effects of
parameters, such as search width, on the output, such as the
segmentation quality.  We then discuss related work and conclude.
Appendix \ref{app:dlr} explores edge weight updating and motivates a
decay schedule for the learning rate, and Appendix \ref{app:bin}
presents experiments where the lowest level input is converted to
binary strings, \ie what happens when we begin with only two concepts,
0 and 1?

%% file: overview.tex
\section{Overview of Expedition}

The input to Expedition is a stream of text, where the input stream is
broken into lines in our experiments, each line of text being the
input for a learning {\bf \em episode}.
We remove blank spaces in the experiments but otherwise we do not do
any extra processing. Thus the line "An apple (or 2) a day!" is input
to the system as "Anapple(or2)aday!".  One reason we remove blank
spaces to see whether and how well the system can learn words and
phrases without the aid of separators.  The only concepts in the
system correspond to characters in the beginning, which we refer to as
{\bf \em primitive} concepts. There is a one-to-one correspondence
between the primitive concepts and characters. A concept corresponds
to a string or a consecutive sequence of characters, but is more than
just a representation however, as it maintains a rich set of
connection information, such as information about its associations
(prediction edges) as well part-related connections
(Sect. \ref{sec:cons}).

\subsection{Episodic Tasks: Segmenting and Updating Weights}

The system is repeatedly fed a line of text, which is readily
converted to a sequence of primitive concepts.  The system then
segments this buffer into a small subset of its (current) higher-level
concepts.  The concepts in the final segmentation are the {\bf \em active}
concepts. These include the highest level concepts the system 'sees'
in an episode. The system updates various statistics, such as concept
co-occurrences, and prediction weights among the active concepts. In
our implementation, these statistics are kept with the concepts
themselves (Section \ref{sec:cons}). These learned statistics
influence future segmentations by the system, and the segmentations
determine what (higher-level) concepts are seen in the input, and thus
which co-occurrences are updated, which in turn affects which
subsequent higher level concepts are ultimately discovered.



\subsection{Periodic Phase: Misc. Tasks Such as Adding New Concepts}
Periodically, such as every say 1000 episodes, the system
performs the {\em periodic functions}, such as updating concept
priors, and new concept constructions, via composition, using
co-occurrence statistics or the prediction weights.  The system can
also occasionally add a new layer of concepts (explained next in
Sec. \ref{sec:layers}).



Initially, the segmentation is trivial or basically given, as each
episode is a concatenation of characters, meaning the input buffer is
a sequence of the corresponding primitive concepts.  Over time, larger
concepts are built, by the composition operation, and segmentation
becomes a non-trivial search problem. The composition operation in our
implemented system is a binary operation, putting two concepts that
co-occur sufficiently frequently together (Sect. \ref{sec:comp}).

\subsection{Levels, or Layering of Concepts}
\label{sec:layers}


A concept corresponding to a low level string, such as a single
character, for example 'a', occurs with other low level concepts, \ie
other characters, but after some learning, it can co-occur with higher
level concepts too. For example, 'a' can co-occur with a concept
corresponding to 'b' and also with a concept corresponding to 'book'.
To the extent possible, we do not want to mix (or confuse)
co-occurrence statistics for the concept (corresponding to) 'a', \ie
when it occurs with primitives in the lowest layer \vs when it occurs
with higher level concepts. Thus, we have found it useful to
differentiate and support several {\bf \em levels} or a {\bf \em
  layering} of concepts that we describe next.  We will use the terms
level and layer interchangeably. Each level has the potential to learn
a more powerful (conditional) distribution of the input.





Layer zero contains the primitives (the primitive concepts) and only
the primitives, and initially only layer 0 exists. Layer $i$, $i\ge
1$, is created after sufficient experience with concepts at layer
$i-1$.  Each level $i \ge 1$, has a {\bf \em clone} or a replica of
each of the concepts in the lower layer, $i-1$, as well as the newly
created bigrams, or {\bf \em holonyms}, of concepts from layer
$i-1$. Therefore, a concept in level $i \ge 1$ has up to (a maximum
of) $i$ composition (concatenation) operations. Crucially, even when a
concept in layer $i$ is a clone of a concept in layer $i-1$, its
co-occurrence and prediction connections are only to layer $i$
concepts and thus they are entirely different (\eg the concept
(corresponding to) 'a' in layer 0 has a different set of prediction
weights from its clone, the concept 'a' in layer 1, etc.). This
(separation of prediction edges) is the original motivation for the
idea of layering. Figure \ref{fig:example_edges} shows a concept with
several connection examples. We also note that in this implementation,
concepts, in particular the clones, become more {\em specific} as we
go up the levels (see Sec. \ref{sec:spec}). The segmentation process
needs to support turning concepts in the input buffer from layer $i$
to layer $i+1$, all the way to the current highest layer.


Figure \ref{fig:codeoverall1} shows the general learning loop.  We
flesh out the important functions in the following sections.

\begin{figure}[ht]
\begin{minipage}[t]{0.45\linewidth}
\centering
  \begin{pseudo}[kw]*
    \hd{\bf{MainLearningLoop}}(T, k) \\*
    & \ct{T is a text corpus/stream, k is the number to sample} \\*
    & \ct{from and train on before each periodic task.}  \\*
    & Repeat \\+*
    & \ct{Predict and segment, do weight updates. } \\*
    & TrainOnEpisodes($T, k$) \\*
    & \ct{Create new concepts, possibly add a layer, etc. } \\*
    & DoPeriodicTasks() \\-*
  \end{pseudo}
\caption{The main loop: Repeatedly perform the episodic leaning. And
  once in a while, do the periodic tasks.}
\label{fig:codeoverall1}
\end{minipage}
\hspace{0.5cm}
\begin{minipage}[t]{0.48\linewidth}
\centering
  \begin{pseudo}[kw]*
    \hd{\bf{TrainOnEpisodes}}(T, k) \  \\*
    & \ct{Repeatedly take an episode and learn from it.} \\*
    & Repeat for k iterations: \\+*
    & \ct{Sample a line (episode e) from stream T} \\*
    & $e \leftarrow$ SampleLine($T$) \\*
    & \ct{Segment and pick a segmentation chain $sc$.}\\*
    & $sc \leftarrow$ Segment($e$) \\*
    & \ct{Update weights \& misc statistics given }\\*
    & \ct{the (segmentation) chain $sc$.} \\*
    & UpdateActiveConcepts($sc$) \\*
  \end{pseudo}
  \caption{The learning inside each episode involves segmenting
    first.}
\label{fig:codeoverall2}
\end{minipage}
\label{fig:codeoverall}
\end{figure}

%% file: concepts.tex
\section{The Network Structure: Concepts and Edges }
\label{sec:cons}

In this work, a {\bf \em concept} corresponds to a string of one or more
characters.
Initially, the system only contains the {\bf \em primitive}
concepts, each primitive corresponding to a single
character.  Over time the system acquires higher level concepts,
corresponding to larger strings by concatenating (composing) lower
level concepts. The system generates a tiny subset of all possible
strings of length $k$ (a diminishing fraction of all possible strings,
as $k$ grows), \ie those that are (likely) meaningful.  To do this,
each concept keeps a number of connections, the 'horizontal' or left
and right connections (prediction weights and co-occurrences), as well
as the 'vertical' or up and down edges (parts and part-of
connections). These connections allow for predictions and are also
used for segmentation.  We next go over the notation we use to refer
to concepts and then describe in further detail what each concept
keeps track of.

\subsection{A Notation for Concepts With Some Examples}
\label{sec:cnote}

We denote the concept corresponding to a string $x$ in level $i$, when
the concept exists, by $\con_i(x)$, for example $\con_3('ther')$. This
notation is specially useful when explaining segmentation for multiple
layers in Sect. \ref{sec:seg_struct}. In layer 0, every character seen
in input has a corresponding concept. In our implementation, a new
primitive concept is allocated in layer 0 the first time a character
is seen (and if there already exist higher layers, the appropriate
clones for that character are generated too). Thus we can have
$\con_0($'r'$)$, and once layer 1 is created, we'll have $\con_1($'r'$)$,
where $\con_1($'r'$)$ is the (unique) clone of $\con_0($'r'$)$.  The
prediction connections and co-occurrence statistics of $\con_1($'r'$)$
involve only concepts in layer 1, while its part-of connections go to
layer 2, and its parts connections, if any (none for $\con_1($'r'$)$,
since it's a clone), go to layer 0. Similarly, $\con_2($'r'$)$ is the
clone of $\con_1($'r'$)$.

We note that $\con_0($'re'$)$ is always undefined or meaningless,
since only single character concepts, the primitives, exist in layer
0. More generally, no concept corresponding to $k > 1$ or more
character-long string can exist in a layer below level $\log k$ due to
our binary composition constraints.

\subsection{Associative Connections ("Horizontal")}
\label{sec:edges}

\begin{figure}[!htbp]
\begin{center}
  \centering
  \subfloat[  ]{{\includegraphics[height=6cm,width=12cm]{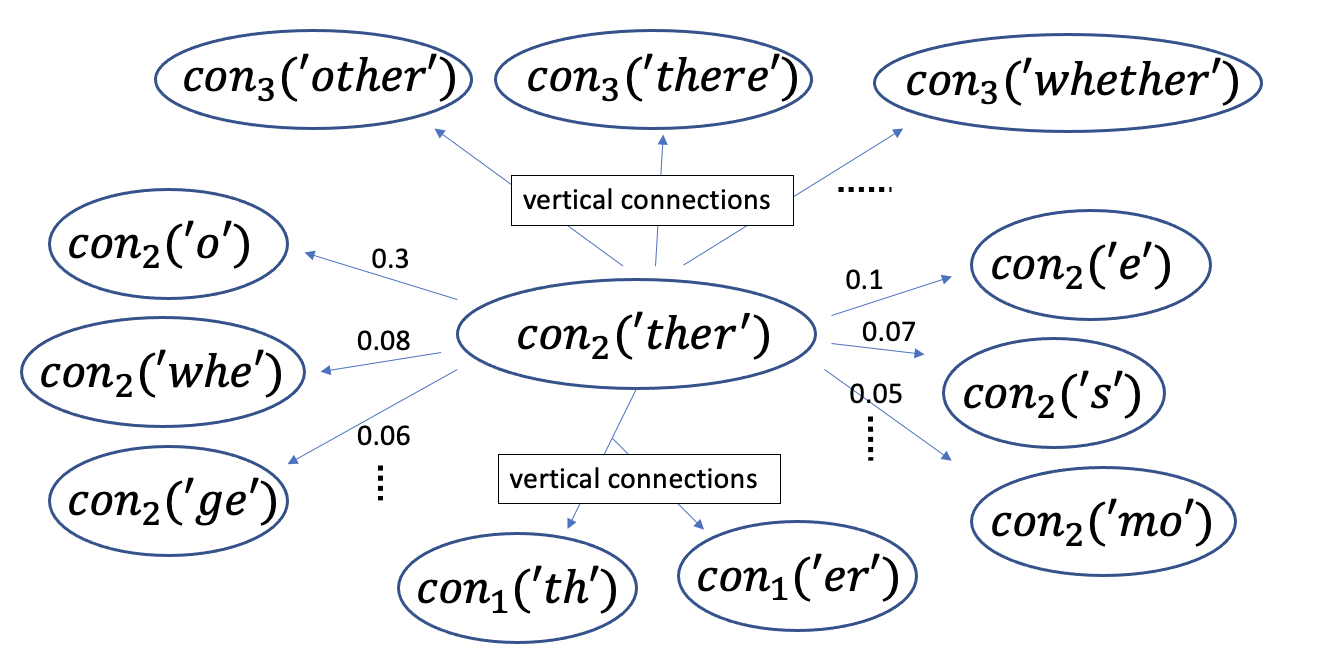} }}
\end{center}
\vspace{.2cm}
\caption{A concept, $con_2('$ther$')$ ("ther" at level 2), from
  a model trained up to level 3, and a few of its horizontal (three at -1
  to the left, and three at +1 to the right) and vertical edges (top 3 most
  frequent of its holonyms in level 3, having $con_2('$ther$')$ as a part). "o" and
  "whe" occur immediately before "ther" with high probability (0.3 and
  0.08 respectively). Table \ref{tab:top10} contains additional
  examples of learnt prediction edges.}
\label{fig:example_edges}
\end{figure}

Each concept keeps and updates edges with weights, the {\bf \em
  prediction edges} or the 'horizontal' connections (imagining text
is written/read horizontally), to other concepts that occur with it,
in segmentations, within a window of size at most some $k$ concepts in
our experiments. In the implementation of this paper, $k$ is set to 3
and the left and right occurrences are distinguished.\footnote{More
  generally, one could also keep connections that are position or
  direction insensitive, but we have not experimented with such.}
These weights are implemented via hashmaps, and hard or soft
constraints on the size of the hashmaps are imposed so the memory
consumption is kept in check and the processing is efficient (\eg
predicting and updating) \cite{updateskdd08}.  Let $\Delta$ denote the
set of relative positions. Our {\bf \em context size} is 3 in the
experiments, thus the set of (relative) positions is $\Delta = \{\pm
1, \pm 2, \pm 3\}$ ($|\Delta|=6$ possibilities). Each concept keeps a
separate weight map for each position $i \in \Delta$. Example of
(horizontal) edges for positions -1 and 1 is shown in
Fig. \ref{fig:example_edges}.

We will next describe a few properties and semantics of the
weights. Let $w_{c_1, c_2, i}$ denote the weight of prediction edge
from $c_1$ to $c_2$ for position $i\in \Delta$.  The weights are
non-negative, and absence of an edge means 0 weight.  When a concept
is first seen, it has no edges (empty maps).  The system begins with
basically a tabula rasa.  The weights are updated using {\bf \em
  exponentiated moving average (EMA)} updates in our experiments
\cite{updateskdd08}, shown in Fig. \ref{fig:updateconcepts}.  One can
verify that the weights remain in $[0, 1]$ and with the manner of
updating, for a specific position sum to no more than 1.0 ($\forall c,
i, \sum_j w_{c, c_j, i}\le 1$, or the weights of a position, at any
time point, form a semi- or sub-distribution).  Moreover, under fairly
general assumptions (\eg taking into account the learning rate $r$ of
EMA and the budget on number of edges), for each position, the weights
converge to approximate conditional probabilities, \ie for instance
the weight $w_{c_1,c2,1}$ converges to the probability of observing
$c_2$ immediately in the next position, given $c_1$ is observed in
current position.  See Appendix \ref{app:dlr} for a review of
additional properties of EMA.

\subsubsection{Sparse EMA and Rate Decay} Note that the weight updates, which are linear operations, can be
carried out relatively efficiently if the size of corresponding data
structures (\eg hashmaps) are kept in check.  Letting the degree be
$d$ (size of a map), then a weight update takes $O(d)$, while the
totality of concepts can be orders of magnitude larger than $d$.  When
a concept $c_1$ is not already connected to target $c_2$, the edge is
added, the weight of the newly added edge being set to the learning
rate $r$ with the EMA update.  The learning rate should be set so that
it is adequately less than the minimum probabilities that we want to
model (\eg a tenth).  For instance, if the minimum probability of
interest\footnote{The minimum probability leading to good performance
  depends on the domain and some experimentation is in general
  required.} is 0.01, we want it to be say 10x lower, \ie 0.001,
otherwise, if the rate is too large, the error in the conditional
probability estimation can be too large. A rate that's too low may
slow down learning substantially.  To speed up convergence, for each
concept $c$ we can start the rate from a high value, and lower it with
each update for that position, or each time the concept $c$ is seen,
to some minimum positive rate $r_{min}$ such as $10^{-3}$ or
$10^{-4}$. Appendix \ref{app:dlr} motivates this {\bf \em frequency-based
  rate decay} schedule.

\subsection{Cross-Layer (or "Vertical") Connections}

Concepts keep track of their part-related edges, edges to their parts,
and edges to the concepts they are part of, which we can imagine as
vertical, in contrast to the horizontal edges. See
Fig. \ref{fig:example_edges}.

Each concept keeps a list of bottom-up connections to compositions in
the next layer (its immediate {\bf \em holonyms}) that it is a part
of, as well as its clone, in the next layer. These connections are
used during (bottom to top) segmentation. The number of such
connections are kept manageable.  We posited that a concept need only
keep 100 to 1000s of such connections.  For instance, while the
character 'a' may be part of tens of thousands of words and phrases
(concepts) in English, the primitive that corresponds to 'a' will be a
part of only 10s to 100s of significant bigrams. Thus, the layering
and significance tests when composing reduces the connection
possibilities.

Similarly, each concept in a layer $i\ge 1$ keeps a list of {\bf \em
  top-down} connections to its part concepts in layer $i-1$. Note that
a concept corresponding to a string of $k$ characters can in principle
be split into two subconcepts (substrings) in $k$ many ways.  However,
many such possibilities will be insignificant in lower layer and will
not be generated. Still, a concept may have more than 2 parts, \eg
$\con_2($'new'$)$ can have the {\bf \em pair of parts}
$(\con_1($'n'$)$, $\con_1($'ew'$))$, as well as $(\con_1($'ne'$)$,
$\con_1($'w'$))$ (the parts will always be {\bf \em paired}).

The top-down connections are used during matching a candidate
composition during (top-down) segmentation.  Note that, in addition to
segmentation, these vertical connections are also useful in
understanding which string pattern a composition concept corresponds
to.

\subsection{Other Information Kept with Concepts}

We also keep a few (scalar) fields with each concept $c$, including
the {\bf \em historical predicted probability} (probability received
when $c$ occurs in a selected segmentation), denoted $c.hpp$, and
$c.freq$, \ie the frequency the number of episodes seen so far, and
various other statistics and counters (first-seen, last-seen, prior,
$\cdots$). A few of these, such as the $c.hpp$, are used in the
algorithms, for example during the segmentation process for guiding
the search towards a good segmentation. Others are for reporting only,
to get insights into the trajectory of the learning.

%% file: seg.tex
\section{Inference: Segmenting \& Predicting}
\label{sec:segs}

The segmentation process ultimately generates a mapping from
stretches of raw characters in the input to internal concepts, and in
that sense it is "interpretation" too. We refer to it simply as
segmentation.  First we formalize what valid segmentations are, then we
present a segmentation algorithm, a beam search, and describe the
scorings of candidate concepts and segmentations that guide the
search.

\subsection{Segmentation Structure}
\label{sec:seg_struct}
The input to the segmentation process at any given layer $i$, is a
{\bf {\em sequence}} of concepts at layer $i$, and the output is also
a sequence of concepts (a {\bf \em segmentation}), at layer $i+1$. A
generic sequence at a layer $i$, is denoted $c_{i,1}c_{i,2}\cdots
c_{i,k_i}$, first index, $i$, referring to the level, the second goes
over the consecutive positions: if the sequence is $k$ concepts long,
then there are $k$ positions. We use the shorthand $\msq{c_{i,j}}$ for
a sequence (at level $i$).




Thus, a segmentation is simply a sequence of concepts in our
implementation.  An example, segmentation of the string "new", is
described below.  In an episode, given is an input character sequence
$\msq{c_j}$, $\msq{c_j}=c_1c_2\cdots c_{k_0}$, of length $k_0 \ge 1$,
which is readily converted to the corresponding primitives concept
sequence in layer $0$, $\con_0(c_1)\cdots
\con_0(c_{k_0})=\msq{c_{0,j}}, 1\le j\le k_0$ (with $k_0$ positions).
Segmentation of this sequence (into concepts of next layer) yields a
concept sequence at layer $1, \msq{c_{1,j}}, 1\le j\le k_1$, where
$k_1 \le k_0$. This process is repeated until we get a segmentation,
\ie a concept sequence, at the highest layer.  Thus the output of
segmentation from one layer forms the input sequence for the next
segmentation process.  The constraint that ties the segmentations
across layers, in this work, is that the concept sequence in layer $i,
i \ge 1$, must {\em exhaustively and partitionally cover} the concept
sequence at layer $i-1$. Next we describe what we mean by {\bf \em
  covering}.

In our implementation, each concept in layer $i+1$ is either a clone
of a layer $i$ concept or is composed of two parts, \ie it is the
concatenation of two (consecutive) layer $i$ concepts. Thus there are
two possibilities for (a valid) covering: 1) Concept $c_{i+1,j}$
covers the single position $j'$, or {\bf {\em matches}} the concept at
that position, in the segmentation for layer $i$ iff $c_{i+1,j}$ is
the clone of $c_{i,j'}$ (the concept in position $j'$), or 2)
$c_{i+1,j}$ covers the consecutive positions $j'$ and $j'+1$, or {\em
  matches} $c_{i,j'}$ and $c_{i,j'+1}$, iff $c_{i,j'}$ and
$c_{i,j'+1}$ are paired parts of $c_{i+1,j}$.

A concept sequence in layer $i+1$,
$\msq{c_{i+1,j}}=c_{i+1,1}c_{i+1,2}\cdots c_{i+1,k_{i+1}}$ is a
segmentation of the concept sequence in layer $i$ iff every concept in
layer $i+1$ covers one or two positions in layer $i$ as defined above,
with the following additional constraints: 1) consecutive concepts
cover consecutive and non-overlapping positions, \ie if $c_{i+1,j}$
covers up to position $j'$ (of layer $i$ segmentation), then
$c_{i+1,j+1}$ covers starting from position $j'+1$, and 2) all
concepts in layer $i$ are covered (by exactly one concept in
the $i+1$ sequence).


As an example, if the input character sequence is $c_1=$'n',
$c_2=$'e', and $c_3=$'w', thus $k_0=3$, then the primitives sequence
is $\msq{c_{0,j}}=[\con_0($'n'$), \con_0($'e'$), \con_0($'w'$)]$ and
we have 3 positions to cover. A (valid) segmentation in layer 1 would
be $\msq{c_{1,j}}=[\con_1($'ne'$), \con_1($'w'$)]$, where $\con_1($'ne'$)$
covers positions 0 and 1 in layer 0, or matches $c_{0,1}$ and
$c_{0,2}$, and $c_{1,2}=\con_1($'w'$)$ is a clone of (and matches)
$c_{0,3}$ and covers position 2. This segmentation is partitional, \ie
the same position in layer 1 is not covered by more than one concept
in layer 2, and exhaustive or complete, in that every position in
layer 1 is covered.  A segmentation at layer 2 would be
$[\con_2($'new'$)]$, covering positions 0 and 1 in the previous
layer. Another segmentation in layer 2 would be $[\con_2($'ne'$),
  \con_2($'w'$)]$ (use clones to cover both positions).

In the future, we may want to relax either conditions of partitionality
or completeness, \eg allow segmentations that overlap to some extent
and/or do not accurately cover every lower level position, allowing
some mismatches or approximate matches, possibly with some
penalization for mismatches, in order to train robust models that can
handle noise/corruption and occlusion in the input.

\subsection{The Segmentation Algorithm}

In the segmentation algorithm presented here, segmentation proceeds
one layer at a time, a segmentation at layer $i$, yields one or more
candidate (valid) segmentations at layer $i+1$.  Figure
\ref{fig:segment} shows the segmentation algorithm.  The process of
segmenting a layer $i$ sequence to get one or more candidate
segmentations at layer $i+1$ is the same for all layers $i$, and we
explain the process next.

Briefly at a high level, segmentation (search) goes as follows.  Given
a layer $i$ sequence, the input layer, initially all its concepts
(positions) are marked uncovered. Pick a remaining uncovered concept
at random or by some quality score (Sections \ref{sec:scoring1} and
thereafter). The quality score could be the maximum score over the
concept's holonyms in the next layer. The concept's holonyms, as well
as its clone, are then matched against the buffer.  The clone always
matches, and a few holonyms may match too.  One of the matches is
picked, either purely at random or by a function of the concept
quality score (see Sect. \ref{sec:localmoves}). The matched one or two
concepts in the input layer (layer $i$) are marked covered, and we
repeat the process for remaining uncovered concepts until all are
marked covered. Once all are covered, we have a candidate segmentation
at layer $i+1$.

Not all segmentations are equal, for instance, a good segmentation for
``anewbike'' is, ``a'', ``new'', ``bike'', but if the system initially
joins ``a'' and ``n'' together to get the concept ``an'', and commits
to it, we get a poor segmentation for the rest of the string "ewbook".
The system performs a beam search to pick the most promising
segmentation at the highest levels. In order to select the most
promising at a given layer and guide the search, it assigns a score to
each candidate segmentation generated.  Scoring a segmentation is in
turn a function of the concepts in the sequence, in particular how
well the concepts in the segmentation "fit" or cohere with one
another, or what we may refer to as {\em coherence}, as well as how
long the concepts are, as we want the system to learn longer
concepts. A third factor is how well a concept matches the input, but
in this implementation we assume all matches are perfect (a match is
all or nothing).  The handling of noise and approximate matching,
leading to a more relaxed and potentially more powerful segmentation,
is an important direction that we leave to future work.


The scoring of the segmentation will also be used as a measure of the
(learning) progress of the system. Table \ref{tab:seg_examples} in the
experiments section shows a few example segmentations.

\begin{figure}[htb]
\begin{minipage}[t]{0.55\linewidth}
{\bf Segment}($e$) // Segment episode (line) $e$. \\
    \hspace*{0.3cm} // Convert $e$ to corresponding primitives sequence: $segs_0$ \\
    \hspace*{0.3cm} // is a list containing a single, layer 0, segmentation. \\ 
    \hspace*{0.3cm} $segs_0$ $\leftarrow$ {\bf ConvertToPrimitives}(e)  \\
    \hspace*{0.3cm} // In each itr., given $segs_{i-1}$, make next layer sequences  \\
    \hspace*{0.3cm} // (candidates), score each, pick top $k$, to get $segs_{i}$.   \\
    \hspace*{0.3cm} Repeat from layer $i=1$ to $l_{max}$: \\
    \hspace*{0.6cm} $segs_i \leftarrow$ {\bf GetNextLayerCandidates}($segs_{i-1}$)    \\
    \hspace*{0.6cm} {\bf FastScoreKeepBest}($segs_i$) // score each, keep best $k$. \\
    \hspace*{0.3cm} // Take highest scoring segm. at top layer and its chain. \\
    \hspace*{0.3cm} Return {\bf GetBestChain}($sc$) \\ \\
            {\bf GetNextLayerCandidates}($segs_i$) \\
    \hspace*{0.3cm} // From layer i to layer i+1 segmentation candidates.\\
    \hspace*{0.3cm} $segs_{i+1} \leftarrow [ \ ]$ // Empty list.  \\
    \hspace*{0.3cm} Repeat $k$ times for each $seg \in segs_{i}$: \\
    \hspace*{0.6cm} $segs_{i+1}\leftarrow segs_{i+1}\cup$ {\bf Generate\_A\_Segmentation}($seg$)\\
    \hspace*{0.3cm} Return $segs_{i+1}$ \\ \\
  {\bf Generate\_A\_Segmentation}($s_i$)  \\
  \hspace*{0.3cm} // Mark all positions in $s_i$ as uncovered. \\
  \hspace*{0.3cm} {\bf MarkUncovered}($s_i$)  \\
  \hspace*{0.3cm} $s_{i+1} \leftarrow [ \ ]$ // empty sequence. \\
  \hspace*{0.3cm} Repeat until all positions in $s_i$ are covered: \\
  \hspace*{0.6cm} $j \leftarrow$ {\bf PickUncovered}($s_i)$ // Randomly. \\
  \hspace*{0.6cm} // Pick a match, update covered, and grow $s_{i+1}.$  \\
  \hspace*{0.6cm} {\bf SegmentationMove}($j, s_{i}, s_{i+1}$)  \\  
  \hspace*{0.3cm} Return $s_{i+1}$ \\  \\
\end{minipage}
\hspace{0.3cm}
\begin{minipage}[t]{0.45\linewidth}
  {\bf SegmentationMove}($j, s_{i}, s_{i+1}$) \\
  \hspace*{0.3cm} $matches \leftarrow$ \{(clone of $s_i[j]$ in layer $i+1$, $j$, 0) \} \\ 
  \hspace*{0.3cm} // Try holonyms for a left and right match. \\
  \hspace*{0.3cm} If position $j+1 < |s_i|$ and uncovered in $s_i$: \\
  \hspace*{0.6cm} $matches \leftarrow matches \ \ \cup $ \\
  \hspace*{0.8cm} {\bf FindMatchingHololnym}($c_j, s_i[j+1], +1$) \\
  \hspace*{0.3cm} If position $j-1 \ge 0$ and uncovered in $s_i$:  \\
  \hspace*{0.6cm} $matches \leftarrow matches \ \ \cup $ \\
  \hspace*{0.8cm} {\bf FindMatchingHololnym}($c_j, s_i[j-1], -1$) \\
  \hspace*{0.3cm} {\bf PickMatchAndUpdate}($matches, s_i, s_{i+1}$)  \\ \\
  {\bf FindMatchingHololnym}($c_1, c_2, pos$) \\
  \hspace*{0.3cm} // $pos \in \{-1, +1\}$. \\
  \hspace*{0.3cm} For each holonym $c_3$ of $c_1$:\\
  \hspace*{0.6cm} If ($pos=+1$ and $c_3 = c_1c_2$) or\\
  \hspace*{0.9cm}  ($pos=-1$ and $c_3 = c_2c_1$):\\
  \hspace*{0.9cm} Return $\{(c_3, j, pos)\}$ // At most one match. \\
  \hspace*{0.3cm} Return $\{\}$ // None found. \\ \\ 
  {\bf PickMatchAndUpdate}($matches, s_i, s_{i+1}$) \\
  \hspace*{0.3cm} // Pick a match, update covered positions in $s_i$, \\
  \hspace*{0.3cm} // and grow $s_{i+1}$. \\
  \hspace*{0.3cm} // Use historical (average) score for selecting. \\ 
  \hspace*{0.3cm} $(c, j, pos) \leftarrow$ {\bf PickBest}($matches$)  \\
  \hspace*{0.3cm} {\bf MarkCovered}($s_i, j$) \\
  \hspace*{0.3cm} {\bf MarkCovered}($s_i, j+pos$) \\
  \hspace*{0.3cm} $s_{i+1} \leftarrow s_{i+1} \cup \{(c, j)\}$ \\
\end{minipage}
\caption{Pseudocode for segmenting an episode, via a beam search. The
  segmentation datastructures, $s_{i}$ (at level $i$) and $s_{i+1}$
  primarily contain a sequence of concepts (initially empty) as well
  as auxiliary fields, such as boolean flag, whether covered, for each
  position. }
\label{fig:segment}
\end{figure}

\begin{figure}[htb]
\begin{minipage}[t]{0.5\linewidth}
    {\bf Coma}($s, optimistic$) \mbox{// Slow segmentation score of $s$.} \\
    \hspace*{0.3cm}    $coma \leftarrow 0$ \\
    \hspace*{0.3cm} // Get probabilities in a take-each-out manner. \\
    \hspace*{0.3cm} for each position $i$ in segmentation $s$: \\
    \hspace*{0.6cm}    $p\leftarrow$ {\bf GetProb}($s$, $i$, optimistic) \\
    \hspace*{0.6cm}    $coma \leftarrow coma$ + $\log(p)$ + {\bf MatchReward}($s[i]$) \\
    \hspace*{0.3cm} Return  $\frac{coma}{|s|}$ \\ \\
    {\bf FastComa}($s$) \mbox{// Uses historical means (concept scores).}  \\
    \hspace*{0.3cm} // Average those scores in segmentation (sequence) $s$.  \\
    \hspace*{0.3cm} Return  $\frac{1}{|s|} \sum_{c \in s}$ {\bf HistoricalComa}($c$) \\ \\
    {\bf GetProb}($s, i, optimistic$)  \\
    \hspace*{0.3cm}    // Return probability of (predicted for) concept $s[i]$. \\
    \hspace*{0.3cm}    if $optimistic$ and $s[i].freq < t_{opt}$:\\
    \hspace*{0.6cm}    Return $1.0$ // Promotes exploration.\\
    \hspace*{0.3cm}    else: // If not in map, Return tiny $\eps$.\\
    \hspace*{0.6cm}    Return {\bf Predict}($s$, $i$).get($s[i], \eps$) \\
\end{minipage}
\begin{minipage}[t]{0.5\linewidth}
  {\bf HistoricalComa}($c$) \\
  \hspace*{0.3cm}  // hpp is historical prediction probability of $c$.\\
  \hspace*{0.3cm}  Return $\log(c.hpp)$ + {\bf MatchReward}($c$) \\ \\
  {\bf{Predict}}($seg, i$) // Predict for position $i$.  \\
  \hspace*{0.3cm} $pmap \leftarrow \{\} $ // Empty predictions map.   \\
  \hspace*{0.3cm} for $j \in \Delta$: \\
  \hspace*{0.6cm} $pmap \leftarrow pmap$ + weights($seg[i+j], j$) \\
  \hspace*{0.3cm} Return {\bf ToProbabilities}($pmap$)\\ \\
  {\bf{ToProbabilities}}($pmap$)   \\
  \hspace*{0.3cm} // Normalize all weights by the sum.  \\
  \hspace*{0.3cm} if $pmap$ == \{\}: Return \{\} // Return if empty. \\  
  \hspace*{0.3cm} $s \leftarrow \sum_{w\in pmap} w $  \\
  \hspace*{0.3cm} $\forall w_i \in pmap, w_i \leftarrow w_i / s $  \\
  \hspace*{0.3cm} Return $pmap$ // Return the normalized map. \\
\end{minipage}
\caption{Scoring and prediction used for segmenting. $t_{opt}=50$ in our experiments.}
\label{fig:scoring}
\end{figure}

\subsubsection{Scoring Concepts} 
\label{sec:scoring1}
Ideally, we want a smooth measure that improves as the system
constructs larger and larger concepts.  We will use the measure to
guide segmentation, as an objective, and we also report it as one
reflection of the overall progress of the system.  Measures such as
average concept length (number of characters) segmented can be
brittle, and also insufficiently sensitive to the steady but small
progress in prediction.  Perplexity (or equivalently entropy) is
widely used in language modeling \cite{perpl,slm} but perplexity goes
down in general with larger vocabularies, and requires extension to
handle vocabularies where multiple terms (\eg "b", "ba", "bat" and
"bath") can occupy the same location of the input.  Probability loss
measures such as quadratic loss are smooth too, but also decrease
(degrade) in general the more items to predict with. They do not
appear to be suitable as measures of "progress". In the experiments,
we do report on several of these measures as well (quadratic loss,
average concept length, etc.).  But for guiding the segmentation
search, we seek a score that improves as the system expands its
vocabulary of concepts. In this and next section, we develop a measure
that improves (increases) in general as the number and extent of
concepts grow over time.


We will be using in part how each candidate concept "fits" with others
in a candidate segmentation, and our measure for this fitness is how
well a concept is predicted, \ie the probability that it attains, from
the local context (the predicting concepts within $\Delta$ positions).
However, a major challenge in scoring a segmentation is that the
concepts can be relatively new, and in general, a segmentation will
contain concepts with widely different frequencies or occurrence
counts. We posited that a concept needs to be seen 100s of time before
its own weights and probabilities to it from concepts that it
co-occurs with begin converging to a stable range.\footnote{Due to
  sparsity and long tails, legitimate co-occurrences may not occur
  even after 100s of episodes of experience with a concept. See
  Sec. \ref{sec:prob_discuss} for a future direction for handling
  sparsity.}  Some period of exploration is required for (relatively)
new concepts, achieved via some regime of randomization as well as
other exploration techniques. We describe how we promote exploration
below.




%
%

Formally, we define the {\bf \em match (intrinsic) reward} of a concept
$c=[c_i]$ as:

\begin{align}
  \match(c) = -\log(\prod_{1\le i\le k}{\prior(c_i)}) =
  - \sum_i{\log(\prior(c_i))} \mbox{, where $c=c_1\cdots c_k$ ($k\ge 1$.)} 
\end{align}

Thus, longer concepts (concepts with more primitives) and concepts
with more infrequent primitives have higher intrinsic reward.  The
priors of the primitive (their occurrence probability) are updated in
the periodic tasks. In a segmentation, the intrinsic or matching
reward of a concept $c$ is balanced against how much probability the
rest of the segmentation (the concepts in the context) assigns to $c$
(the coherence part), to get a COherence+MAtch,\footnote{Another way
  to look at it is that it is a balance between horizontal or
  side-ways fit \vs vertical or top-down fit (match). } or {\bf \em
  \coma} {\bf \em concept score}, as follows:

\begin{align}
  \score(c) = \log \frac{\pred(c)}{ \prod_i{\prior(c_i)} } = \log
  \frac{\pred(c)}{-\match(c)} = \match(c) + \log(\pred(c)) \
  \mbox{({\bf \em concept (\coma) score})},
  \label{eq:cscore}
\end{align}

where $\pred(c)$ denotes the (prediction) probability assigned to $c$
by the context concepts in the segmentation (note: $\log(\pred(c)) \le
0$).  See GetProb() and Predict() in Fig. \ref{fig:scoring}.  To
motivate this set up, we imagine comparing the prediction system
against a baseline system that predicts at the character level (never
learns bigger patterns). The baseline makes the independence
assumption when predicting and does not use any context. Therefore, it
assigns $\prod_i{\prior(c_i)}$ to a concept $c=[c_i]$ (irrespective of
the segmentation).  The log of the ratio,
\begin{align*}
\log(\frac{\pred(c)}{\prod_i{\prior(c_i)} }), 
\end{align*}
of the probabilities assigned by Expedition and the baseline systems
to $c$, is the score or the reward of the system for predicting
concept $c$ with probability $\pred(c)$. The farther the system gets
from the baseline in the above sense (\ie the larger the ratio), the
higher the system's score.\footnote{Note that the score can be
  negative when the ratio is below 1, and this does happen in our
  experiments, even on an average basis for some concepts. For
  instance, the concept may be inferior, or due to the poor quality of
  the probabilities computed by the system, or the difficulty of the
  task.  For example, see Appendix \ref{app:bin}, where negative
  scores are (initially) prevalent. }

Another way to see how the above objective promotes using composition
(larger) concepts: consider whether to join concept $c_1$, that is
predicted on average with probability $p_1$, with a primitive concept
$c_2$ with prior $p_2$. Say the joining, or the composition $c_1c_2$
would be predicted on average with probability $p_3$. Then the joining
is beneficial (increases the score on average) iff $\frac{p_3}{p_1} >
p_2$, \ie even though in general we have $p_3 < p_2$ (the probability
derived from prediction of $c_1c_2$ in one shot will be in general
less than probability of $p_1$ of just predicting $c_1$), as long as
the reduction is not more than $p_2$, it pays to join the two and
predict them together.




\subsubsection{Promoting Exploration}

For any concept $c$, the prediction probability $\pred(c)$ derived
from the predictors in the context, will take time to learn and
requires using ('seeing') concept $c$ in segmentations. To promote
exploration, \ie the use of newly generated concepts or concepts that
are relatively infrequent, a concept that has been seen less than $t$
times so far (in any segmentation) gets a probability of 1.0 ({\bf
  \em optimistic probability}), irrespective of the actual probability
it receives by the context (the predictor concepts nearby):

\begin{align}
  \label{eq:optprop}
  \pred_{opt}(c) &=& \hspace*{-1in} \begin{cases}
    1.0 \mbox{\ \ if \ } c.freq \le t_{opt}  \mbox{\ \ ($t_{opt}=50$ in our experiments.)}\\
    \pred(c) \mbox{\ \ otherwise (via aggregating predictions from context).}
  \end{cases} \\
  \score_{opt}(c) &=& \match(c) + \log(\pred_{opt}(c)) \mbox{\
    ({\bf \em Optimistic concept score (optimistic \comap)})}
\end{align}

The concepts in the context update their prediction weights to a
target concept in the usual manner. Note that when $\pred_{opt}=1.0$,
$\score_{opt}(c) = \match(c)$.  We have experimented with thresholds
$t=20$ and $t=50$, and $t=50$ appears to provide adequate time for
learning for our experiments, but a longer period (bigger $t_{opt}$)
may be needed in practice. After this period, the moving average of
the actual (predicted) probabilities are used.

\subsubsection{Using Historical Averages of Concept Scores}

To make local segmentation moves, \ie selecting among matching
concepts at the next layer, which eventually creates a complete
segmentation candidate at the next layer (see SegmentationMove() in
Fig. \ref{fig:segment} and the next section), as well as to quickly
score an entire candidate segmentation (Sect. \ref{sec:coma}), we use
a {\bf \em historical} version of the \coma score, \ie a combination
of average of probability and $\match()$ of a concept (defined next)
to select among matching alternatives, as there is no full
segmentation yet available.

For keeping an average we use the EMA function.  When a concept is new
($c.freq \le t_{opt}$), its hpp, {\bf \em
  historical-predicted-probability ($c.hpp$)} is 1.0.  Once $c.freq > t_{opt}$
the hpp of the concept $c.hpp$ is updated via EMA whenever the concept is active, \ie it occurs
in a final selected segmentation,  as shown in
Fig. \ref{fig:updateconcepts} (in function UpdateActiveConcepts()) in
Sec. \ref{sec:learning}.  The hpp is thus the moving average of $prob_{opt}$
from Eq. \ref{eq:optprop}.



The historical score of a concept, $\score_{hist}(c)$, is thus:

\begin{align}
  \label{eq:hist}
  \score_{hist}(c) = \match(c) + \log(c.hpp) \mbox{\ \ \ \ ({\bf \em Historical concept score})}
\end{align}

\subsubsection{Beam Search and Making Local Moves}
\label{sec:localmoves}


At each layer, we keep up to $w\ge 1$ candidate segmentations. Each
such concept sequence at layer $i$ is tried $b \ge 1$ times to
generate $b$ candidates (that we describe next), thus we get up to the
product $wb$ segmentation candidates at $i+1$. Some may be duplicates,
so we may get fewer unique segmentations.\footnote{In our
  implementation, we keep one candidate segmentation from those that
  have identical score.} Of these, (up to) width $w$ highest scoring
are kept. The next section describes scoring entire segmentations. The
process is repeated, at each next (higher) layer, until the current
highest layer is reached. Fig. \ref{fig:segment} presents pseudocode
for a few of the search functions.

We will use the terminology and the notation introduced in
Sect. \ref{sec:seg_struct}. To generate a single candidate
segmentation at level $i+1$, $s_{i+1}$, from $s_i$ of level $i$
(function Generate\_A\_Segmentation() in Fig. \ref{fig:segment}),
local matching and covering moves are repeatedly made until all
positions of $s_i$ are covered: a random uncovered position $j$ at
level $i$ is picked, the concept at position $j$ tries its clone and
its holonym concepts (which belong to layer $i+1$) and the match with
the highest historical score ($score_{hist}()$ of Eq. \ref{eq:hist})
is picked.

For example, assume "new" is in the primitives layer represented as
$s_0=[\con_0($'n'$),\con_0($'e'$),\con_0($'w'$)]$, and assume location
2, \ie $\con_0($'w'$)$, is selected, out of the three positions, to be
covered first. Assume the holonyms of $\con_0($'w'$)$ include
$\{\con_1($'we'$), \con_1($'wu'$), \con_1($'ew'$), \con_1($'ow'$),
\cdots\}$. Then we get two matching candidates $\con_1($'ew'$)$ and
$\con_1($'w'$)$, and whichever has higher (historical) score is
picked. Assume $\con_1($'ew'$)$ is picked. Then positions 0 and 1 are
marked covered, and the remaining uncovered position 2, or concept
$\con_0($'n'$)$, is next covered by a clone, yielding
$s_1=[\con_1($'n'$),\con_1($'ew'$)]$.

We make the following notes regarding our current implementation: 1)
The clone always matches (there is always at least one match). 2)
There are at most 3 candidate matches to pick from, one is the clone,
and up to 2 holonyms, one matching to the left (covering positions $j$
and $j-1$) and one to the right of $j$, when those positions exist and
not yet covered (\eg three is no left-matching holonym at $j=0$). 3)
The main source of randomization here is in picking an uncovered
position. We leave sampling from the matches (to promote more
exploration) to future work (see discussions below on interaction of
segmentation and learning). 4) Each picked (matching) concept in layer
$i+1$ can keep track of the position $j$ it matched (one of the up to
2 positions suffice), and once all positions in $i$ are covered, a
sort by $j$ of the picked gives a complete candidate segmentation at
layer $i+1$. Note also that the same concept can match and be picked
for multiple positions. Simple list and array datastructures are used
to implement this search.


\subsubsection{Scoring and Selecting Candidate Segmentations}
\label{sec:coma}

When candidate complete segmentations at a given layer are computed,
we still use the historical averages, with the possible exception of
the top layer, to score candidates and select candidates (the fast
score):

\begin{align}
  \label{eq:fast}
  fast\_coma(s) = \frac{1}{|s|} \sum_{c\in s} \score_{hist}(c)
  \mbox{\ \ \ \ ({\bf \em fast score} of a segmentation, via historical concept scores). }
\end{align}

However, we do report on the actual and optimistic \coma scores of the
selected segmentation from the top layer:

\begin{align}
  \label{eq:coma}
  coma(s) = \frac{1}{|s|} \sum_{c\in s} \score(c) \mbox{
    \ \ ({\bf \em actual, non-optimistic, \comap}, for a segmentation.)} \\
  \label{eq:coma2}
  coma_{opt}(s) = \frac{1}{|s|} \sum_{c\in s} \score_{opt}(c) \mbox{
    \ \ ({\bf \em optimistic \comap}, for a segmentation.) }
\end{align}

The \coma score incorporates the (match) reward of each concept with
the probability it attains from the context (the rest of the
segmentation), using Equation \ref{eq:cscore} for the score of each
concept. However, during the beam search for a good segmentation,
across multiple layers, computing the prediction probability for each
concept in every candidate segmentation can take a long time. This is
a main reason we rely on the fast score.  Another consideration is
that the goal during segmentation across layers is to get a good
segmentation at the top layer, and computing \coma may be unnecessary
for the intermediate layers (any layer that is not the top layer).
For the top layer, we note that the fast choice for scoring a
segmentation is less accurate and uses a concept's average score
instead of its current score in a segmentation. The average score
reflects how well a concept {\em in general} fits within its context
(combined with its match reward) instead of how it fits within its
context in the current episode. For the top layer, we can use the
(optimistic) \comap, and in a main set of experiments we do that (See
Sect. \ref{sec:the_two}). However, we leave the question of the extent
to which such distinctions make a difference on the learning
trajectory of the system to future work. Sect. \ref{sec:exp_segscores}
presents graphs that include the evolution of scores of all the
scoring techniques.

Each segmentation at layer $i\ge 1$ keeps track of which segmentation
at $i-1$ led to it (if there are multiple ones leading to the same,
pick one at random), and we thus get chains of segmentations across
layers.  In the highest layer, the highest scoring segmentation (ties
broken at random) and its chain are kept. The various fields of the
concepts in the selected chain (such as prediction weights and seen
counts), the {\em active concepts} are then updated
(Sect. \ref{sec:learning}).

\subsection{Concept Specificity Increases with Level}
\label{sec:spec} %

Concepts become more {\bf \em specific} as we go up the concept levels in
our current implementation, in the following sense: when the character
"a" occurs in the input, $con_0($'a'$)$ is always activated, \ie it
always occurs in the final selected segmentation, but $con_1($'a'$)$
may not be active, instead a holonym of $con_1($'a'$)$, for instance
$con_1($'an'$)$ may be active, depending on the final selected
segmentation. On the other hand, whenever $con_i($'a'$)$ is active,
$i\ge 1$, $con_{i-1}($'a'$)$ is active too.

\subsection{Special Predictors}
\label{sec:special} %
We implemented three special predictors for each level, the {\em
  begin-buffer}, the {\em end-buffer}, and the {\em always-active}
predictors.  These were added to provide insights (into the input
text), and to possibly improve performance.  The begin-buffer
predictor is a concept that predicts, and updates for, the beginning
of the buffer or the segmentation at that level (the first $3$
positions in our experiments), while the end-buffer does the same for
the last few positions of the segmentation. The always-active predicts
every position, and basically provides a prior for that level.  The
begin and end predictors make it possible to have a well-defined \coma
score at the top level when a single concept may be suitable (\eg
specially for short input lines).

\subsection{Discussion}
\label{sec:inf_discuss} 

There are a number avenues for improving the segmentation process and
making it more powerful, such as exploring other segmentation
objectives, extending the search algorithms and relaxing the matching,
and improving the prediction probabilities.

\subsubsection{Objectives}

Currently, we compare performance with respect to a baseline system
operating at the lowest primitives level. An alternative, for example,
to measure progress, is by comparing prediction performance of each
layer to a baseline that is operating at the layer below. One
advantage of comparing performance to the lowest layer is that this
is the layer that we ultimately care about predicting: predicting
"reality". Higher layers are concoctions of the system itself, and
scores based on comparing to them remain only indirect measures of
performance.

There may also be other baselines to compare to, and our particular
definition of what to compare to and how to compare is improvable. In
particular, the next section discusses relaxing the perfect match
requirement.

\subsubsection{Relaxing the Segmentation}

We want to allow approximate (imperfect) matches to handle the possibility
of additional noise in the input. For instance, imagine typos and
spelling errors. We also want to allow partially overlapping concepts,
and partial (non-complete) segmentations, as a way of handling white
space for example, as well as to better handle cases where the input
is partially hidden.

Currently the search completes a segmentation in one layer, before
proceeding to the next. An interesting alternative which may fit
better with partial and imperfect segmentation is for the search to
"cut-across" layers.  The segmentation \coma objective needs to be
extended to handle such cases too. A general significant challenge is
that, as we extend or relax segmentation and/or extend concept
representation, keeping the (code) complexity and efficiency of
segmentation (inference) in check. This and whether segmentation can
provide the effective feedback for learning more sophisticated concept
structures are open questions.  We also expect that understanding the
interaction of segmentation and the learning trajectory, for example,
how the parameters for one, or ther details of the algorithm for one,
affects the other part, and ultimately the progress of the entire
system, is an important area for future investigation.

\subsubsection{Improving Prediction Probabilities}
\label{sec:prob_discuss}

There exist a number of ideas to improve prediction, and several of
the ideas can be fairly readily implemented within the system.  The
quality of the probabilities assigned are important for scoring
concepts and segmentations (the \coma score).  We do plain normalization
to attain probabilities. We experimented briefly with a plain softmax
function ($\frac{\exp(w)}{\sum \exp(w)}$) but that gave inferior
results in terms of quadratic loss on probabilities, probably because
of too much increase in the probability of those concepts that obtain
higher weights. An adaptive or an online learning of a mapping, an
online variant of binning \cite{zad1}, significantly improves the
predicted probabilities in terms of quadratic loss.  However, we do
not know the extent of the impact of these enhancements on the
learning trajectory.  One can also experiment with weighting the
prediction of different concepts, taking into account the frequency of
a concept and the relative position, \eg via expert weighting methods,
such as \cite{sexperts1}.

Many concepts are of the same type or behave similarly in many
contexts, \eg the to-be verbs, the digits, punctuations, or the upper
and lower case versions of characters. A major direction is ways of
discovering and incorporating such to address sparsity, \ie that is
the problem of zero or very low probabilities for infrequent concepts
and, more generally, unseen or little-seen co-occurrences. There will
always be concepts that the system has limited experience with, and,
with a plethora of concepts, not all legitimate co-occurrences can be
observed with finite experience, and limits of system memory will
always be a hard constraint too.  There is a range of techniques in
statistical language modeling which could be useful in addressing
sparsity.  For instance, analysis and use of the common neighborhoods
in the graph of prediction edges, during inference, can improve
performance in this regard.


%% file: learning.tex
\section{Learning: Updating Weights and Scores, and Composing}
\label{sec:learning}

The learning or updating can be classified into two main forms in
Expedition. One form is updating weights of edges and various
concept-related statistics used to score concepts and candidate
segmentations. These updates take place at the end of each episode
after a final segmentation is selected. Another learning is making new
concepts, or composing.

\begin{figure}
\begin{minipage}[t]{0.5\linewidth}  
    {\bf UpdateActiveConcepts}($sc$) \\
    \hspace*{0.3cm} // Given segmentation chain $sc$, update stats of \\
    \hspace*{0.3cm} // its concepts (prediction weights, occurrence counts..)  \\
    \hspace*{0.3cm} For each segmentation $s$ in chain $sc$: // (for each layer $l$) \\
    \hspace*{0.5cm} $processed \leftarrow \{\}$ //  \\
    \hspace*{0.5cm} For each position $i$ in segmentation $s$: \\
    \hspace*{0.7cm} if $s[i]\not\in processed$: \\
    \hspace*{0.9cm} $s[i].freq$ += 1 // Increment the count seen. \\
    \hspace*{0.9cm} $processed$ += $\{s[i]\}$  // Once per episode. \\
    \hspace*{0.7cm} // update hpp (historical-predicted probability of $c_i$). \\
    \hspace*{0.7cm} $p \leftarrow$ {\bf GetProb}($seq, i, true$) // assigned probability.  \\
    \hspace*{0.7cm} $s[i].hpp \leftarrow$ {\bf EMA}($p, s[i].hpp, r_{mix}$)  \\
    \hspace*{0.7cm} // Update prediction weights. \\
    \hspace*{0.7cm} for each (valid) position $i+j, j \in \Delta$: \\
    \hspace*{1.1cm}    {\bf UpdateForPosition}($s[i+j], s[i], -j$) \\
    \hspace*{0.7cm} // Update other fields of concept $s[i]$, for example: \\
    \hspace*{0.7cm} $s[i].last\_seen \leftarrow$ this\_episode. \\
    \hspace*{0.5cm} // Update other fields, for reporting purposes, etc. \\
    \hspace*{0.5cm} Maintain moving average of {\bf Coherence}($s$, false) \\
    \hspace*{0.5cm} Maintain moving average of {\bf Coherence}($s$, true) \\
\end{minipage}
\begin{minipage}[t]{0.5\linewidth}  
    {\bf{UpdateForPosition}}($c_1, c_2, j$) \\
    \hspace*{0.3cm} // Update prediction weights of $c_1$ for relative position \\
    \hspace*{0.3cm} // $j$. Strengthen weight to $c_2$ (and weaken others).   \\
    \hspace*{0.3cm} $r \leftarrow $ {\bf GetRate}($c_1$) // Learning rate for concept $c_1$. \\
    \hspace*{0.3cm} \ct{Weight map of $c_1$ for position $j$.} \\
    \hspace*{0.3cm} $wmap \leftarrow $ {\bf GetWeights}($c_1, j$) \\
    \hspace*{0.3cm} for all $c \ne c_2$:  \ \ct{Weaken all, except $c_2$'s connection}   \\
    \hspace*{0.6cm} $wmap[c] \leftarrow$ EMA($0$, $wmap[c]$, $r$) \\
    \hspace*{0.3cm} $wmap[c_2] \leftarrow$ EMA($1$, $wmap[c_2]$, $r$) // Boost $c_2$'s weight.\\ \\
    {\bf EMA}($x_{new}, x_{avg}, r$) // Exponential moving average.\\
    \hspace*{0.3cm} // Returns the updated average. \\
    \hspace*{0.3cm} return $(1 - r)x_{avg} + rx_{new}$ // Updated average. \\
\end{minipage}
 \caption{Updating (learning) of prediction weights and historical
   concept scores and statistics ($c.freq$, $c.hpp$, etc.).  In this
   paper, the offsets or relative positions are
   $\Delta=\{\pm{1},\pm{2}, \pm{3}\}$ (up to 3 concept positions left and
   right), $r_{mix}=0.01$.}
 \label{fig:updateconcepts}
\end{figure}

\subsection{Updating}

Figure \ref{fig:updateconcepts} shows the update operations. A number
of updates occur, and all are done on the active concepts once a final
segmentation chain is selected.  The active concepts, \ie the concepts
in each segmentation from every layer, update their individual
statistics, in particular $c.freq$ and $c.hpp$, as well as edge
weights to nearby concepts (for each relative position $i \in
\Delta$).

Appendix \ref{app:dlr} presents several properties of EMA in
particular for weight updating and motivates a {\em frequency-based
  rate decay} schedule.


\subsection{Composition}
\label{sec:comp}

We describe our simple concept generation process, then discuss
alternatives and enhancements.

Each concept keeps statistics about its immediate co-occurrences. In
our implementation, each concept keeps track of co-occurrence counts
on concepts that appear immediately to the right in a segmentation.
Thus, in the final selected segmentation, whenever concept B follows
concept A, the co-occurrence count entry for concept B, in the list of
co-occurrences for concept A, is incremented by 1. If the entry is not
there, it is created.  Similar to other connections, the size of these
lists are kept within a budget.  Also, if a bigram concept has already
been created, the system need not update the co-occurrence counts of
its parts (an efficient lookup check can be performed). We also impose
that either A or B must not be a clone so that duplicate concepts are
not generated. In one set of experiments, we relax this condition, and
we discuss these decisions later below.  Each concept also keeps track
of its overall occurrence counts, from which concept priors can be
computed.


In each periodic phase, the co-location lists are examined, and pairs
of concepts that pass the minimum co-occurrence count (10 in our
experiments) and a minimum binomial tail score of 5 generate new
bigrams, \ie new composition concepts to be used in layer $i+1$. A
tail upper bound of say $x$, bounds the probability to no larger than
$x$, that we observe $k$ or more co-occurrences ($c_2$ following $c_1$
under any model that specifies $P(c_2|c_1)$ is no bigger than the
prior of $c_2$, \ie $\prior(c_2)$)
\cite{binomText2021,omid2020,arratia89,ash90}. The score is the
negative log of the tail bound $x$ ($-\log(x)$), therefore the higher
the (tail) score (the lower the tail bound), the more confident we are
in rejecting that $P(c_2|c_1)$ is any value at or below $\prior(c_2)$
and if we reject, we are concluding the conditional must be larger
(assuming it exists, \ie we assume it is well defined). This check is
a gate for generating candidate holonyms, and compared to pointwise
mutual information, also used for (meaningful) bigram creation
\cite{church90,manning_nlp,exp1}, performs better for discovering
(good) bigrams for frequent concepts
\cite{binomText2021}.\footnote{The approximation to the binomial that
  we use only requires the observed probability $q=p(c_2|c_1)$ and
  $\prior(c_2)$ (the expected proportion by the null model)
  \cite{binomText2021,arratia89}. Therefore, one could use the
  prediction weights of edges, which are approximate moving averages
  of $p(c_2|c_1)$, instead of keeping an explicit list of
  co-occurrence counts (saving time and space).}

There is some possibility that the holonym may already exist, but with
a different pair of parts, and in this case, the new pairs are added
as additional paired parts, so the concept "new" in layer 2 may have
two decompositions in layer 1: "n" and "ew" as well as ``ne'' and
``w''.  We may do holonym generation (and the needed co-occurrence
updates) at the top most layer only, as discussed below.



\subsection{Adding a New Layer}
\label{sec:layer}

In the periodic tasks, once in a while we may also add a new
layer. When adding a new layer the system performs the following.  All
existing concept in top layer $i$ are cloned for new layer $i+1$, and
composition criteria are checked for creating new (non-clone) holonym
concepts for the new layer $i+1$. All concepts at $i+1$ are
appropriately initialized (frequencies initialized to 0, optimistic
historical probabilities at 1, and empty lists of prediction edges and
part-of edges).





In the experiments of this paper, we manually added a layer when
certain criteria of sufficient training or convergence were met.
Specifically, in most experiments, unless otherwise stated, we added a
layer once the (moving) average over several (100s of) episodes of the
minimum frequency of a concept observed in an episode went beyond 500,
meaning that most concepts seen or used in segmentations (see next
section), have reached adequate (100s) of learning episodes.  Other
criteria, that we have experimented with or used (\eg in Appendices
\ref{app:dlr} and \ref{app:bin}) include waiting until optimistic and
actual coherence (Sect. \ref{sec:coma}) converge (\eg to near 10\% of
one another).  Because only a few layers need to be added in general,
we do not think this problem is critical, although how best to
automate it depends on a careful study of how the various choices
affect the (long-term) learning performance, and is worth future
investigation.


\subsection{Discussion}
\label{sec:learning_disc}

An interesting question is whether learning and updating should
continue in lower layers once a new top layer is added, or whether it
should be turned off.  For instance, assume the current top layer is
layer 3. Should learning continue in levels 0 to 2?  There are
tradeoffs of efficiency (learning takes computational resources),
considerations of non-stationarities, and other considerations such as
whether adequate time was spent learning overall, and for the
individual concepts.  This question also depends on the type of
learning. For instance, updating the various types of concept scores
and perhaps prediction weights among concepts may continue in lower
layers, specially if we assume potential non-stationarities, but we
may turn off, at the lower levels, the learning that is for the sole
purpose of creating new concepts.

\subsubsection{Continue Composing At Lower Layers?}
\label{sec:disc_comp1}

Learning new concepts at a lower layer can be problematic because the
same concepts may have already been discovered at higher layers, \ie
duplicate concepts may be generated that then may interfere with one
another. We have observed that such events can occur sufficiently
often and may cause systematic errors if not somehow addressed.  If we
have an efficient process for (near) duplicate detection and handling
duplicate concepts (\eg effective merging), this issue could be
handled well, though we have not pursued this route. As explained
above, in one original set of experiments we do not allow two clones
to compose at a given layer (one concept needs to be a non-clone), but
we allowed concepts composing in lower layers.

However, the above restriction of not composing when both are clones runs the risk
of not creating certain compositions: what if two characters
(primitives) can pass the test for composition, but only in layer 1 or
higher, \ie after segmenting with existing concepts, that is, certain concepts may be discoverable
{\em only after} other
compositions of the same level are discovered and segmented with. We
do not know whether this issue is a sufficiently frequent phenomenon.
In a 2nd set of experiments, we allowed composing at top-most layer
only, but allowed any pair of concepts (clone or not) to compose in
the top layer. This resolves the issue just mentioned, and we do not
run the risk of creating duplicate concepts at different
levels. Furthermore, co-occurrence count updates can be turned off at
lower levels (efficiency, \ie less work in lower levels). However,
this runs the danger of the system getting stuck with (or limited by)
how the segmentation at the lower levels lead it to "see" the world at
higher levels: if there are errors or imperfections with how the lower
levels segment, the system may not be able to recover. For instance,
the input stream may change its nature (nonstionarity), \eg the
language may switch in the text stream.  The system may remain stuck
with how it segments with concepts originally discovered.

Section \ref{sec:the_two} describes the two main settings in our
experiments.  We currently favor the second approach (see Section
\ref{sec:exps_holo}), at least for a non-changing domain and given
that we do not have duplicate removal or merging techniques.  It is
likely that there are inherent tradeoffs and it is impossible to
completely satisfy all the desiderata, but understanding this space
could be a fruitful future direction.

\subsubsection{On Updating Prediction Weights and Concept Scores}
\label{sec:disc_edges}

Updating prediction weights can be relatively slow, as it relies on
'take-each-out' (or leave-one-out) updating. We don't expect the
number of levels to grow to more than 10s say, because the number of
concepts grows at least exponentially in the number of concepts (see
Fig. \ref{fig:cons}), thus updating prediction edges may not be a
large extra cost, at least once a final segmentation chain is
selected, and the updates are done only for such. In the experiments
of this we do this paper, we follow this strategy.

On the other hand, depending on the goals of the segmentation and the
details of the segmentation search, updating prediction weights in
lower layers may not be needed. For instance, one may argue that
reaching a good segmentation at the top layer is the most important
goal, and the task of the concepts at the lower layer is only to
facilitate that search. Then one idea in that direction is that the
lower layer concepts need only keep a running average of the score of
the top layer concept they lead to. The segmentation search can use
those scores for finding good segmentations at the top.

\subsubsection{Criteria for Composition}
\label{sec:disc_comp2}

The criterion of distribution change for concept creation is an
attractive condition for concept generation: \ie only generate
(or keep) a bigram if the distribution sufficiently changes compared
to the distribution of the part concepts. We are basically requiring
sufficient change in "meaning". Otherwise, if using the parts to
predict together, gets close enough to the holonym (the composition),
the composition candidate can be discarded (do not waste further
resources on it).  One can use the prediction edges to determine
distributions.

Questions include which distribution, front or back of the concept, or
a combination of all. Moreover, there will be issues of sparsity and
surface form (similar to document similarity): two distributions may
look different, \ie different words or concepts, but the concepts may
only look different, but have same or similar enough meaning.  The
precise details need to be specified and investigated, \eg what is a
sufficient change in distribution?  and how to do this efficiently:
for instance do we need to generate compositions first and then test,
\ie after sufficient experience decide whether we want to keep a
generated concept?

There may also exist a more direct way of tying concept generation to
the goal of improving the segmentation objective. Currently, we use a
statistical test, the binomial tail, to filter candidates. We believe
the test is a robust way of reducing the number of candidates
(efficiency), and expect that any good concept will pass the test at
some point in practice during learning (\ie with non-adversarial
data), but investigating and establishing this would be useful.  See
also Appendix \ref{app:bin} where the larger concepts can be harder to
discover from a case of the lowest level containing only two
primitives (input strings are binary strings).

%% file: exps.tex
\section{Experiments}
\label{sec:exps} 	%

These experiments are run on the NSF abstracts dataset
\cite{dua:2019}. The dataset contains about 120k abstracts, or 2.5
million English text lines and over twenty million term occurrences.
Each line contains about 55 characters on average. Blank spaces are
removed from each line, to test whether words and phrases are
eventually discovered, and no other processing is done.  Each episode
is a random line of a random abstract.\footnote{The abstracts are
  split into 11 files. The files are randomly permuted in each pass,
  then a random sample of a few 100s of lines are processed from each
  file.}

We report on a number of statistics and measures of progress to get a
sense of how the learning progresses next (Sections \ref{sec:exps_ncons} to
\ref{sec:exps_qloss}). We then give specific examples of concepts,
holonyms, prediction edges, and segmentations, along with several
related statistics, and report on the effects of changing segmentation
search width on segmentation score and the quality of the splits in
the segmentation (Sections \ref{sec:exps_holo} to
\ref{sec:exps_segs}). Sect. \ref{sec:timing} reports on training
times and other computational costs.

\subsection{The Parameters and two Models}
\label{sec:the_two}

In all the experiments, unless otherwise stated, we used a window of
size 3 on both sides of a concept ($|\Delta|=6$) for prediction and
updating, frequency-based decay for the learning rate of the
prediction edge weights (to a minimum rate of $r_{min}=0.0001$), and
budget sizes of 200 for edge-weights (each relative position) and
co-occurrence lists.  We used 10, 3 \ie try 10 and keep 3 for the beam
search of the segmentation algorithm.


We report on two approaches to generating compositions (holonyms). See
\ref{sec:disc_comp1} for a discussion.  In the first approach, we
continued the experiments for some time after layer 4 is created, and
we call the obtained model {\bf Model4}. Here we allowed composition
generation at any level, thus layer 4 concepts and segmentation may be
present but layer 0 can keep creating and adding concepts into layer 1
(at the periodic phase). However, we imposed the condition that at
least one concept in the composition of two needs to be non-clone, so
that we avoid creating redundant concepts at different layers.  In the
second approach, where we continued the experiments for some time
after layer 3 was created, we allow concept creation only at the
highest layer, but any pair of concepts can be composed (both can be
clones). We call the model we obtained {\bf Model3}. In both models,
historical \coma scores were used to pick segmentation moves and
entire segmentations in all layers, except that for Model3, in the top
layer, we used the optimistic \coma score (eq. \ref{eq:coma2}) to pick
a final segmentation in the top-layer (and its chain). In all cases
(whether or not historical), we used the optimistic versions of
\comap.

Model3 is a snapshot at nearly 2 million episodes, and Model4 is a
snapshot at nearly 2.5 million episodes.  The progress in \coma
(segmentation) scores are similar in the two systems. There are a few
differences, \eg in the number of concepts generated or the number of
holonyms of concepts (see Sect. \ref{sec:exps_holo}), and we currently
favor the approach that led to Model3.


\co{ on segmentation at one level higher, \ie increased the number of
  layers by 1, when the minimum frequency of a concept, observed in an
  episode, reached some desired threshold: in our case, in the order
  of 100s (around 500 in these experiments), meaning that most
  concepts observed (see next section), have reached sufficient (100s)
  of learning episodes.

Other criteria we have experimented with
include requiring some minimum $x$ proportion, on average, of concepts
segmented in the highest layer be non-clone concepts (eg $x=0.50$),
although this criterion may not be sufficient as it does not indicate
sufficient training. There could be other criteria, however since the
number of layers won't be very high (in the 10s), we don't expect this
problem to be a crucial.
}

\subsection{Number of Concepts and Their Frequencies}
\label{sec:exps_ncons}

\begin{figure}[!htbp]
\begin{center}
  \centering
  \subfloat[ Number of concepts observed. ]{{\includegraphics[height=6cm,width=8cm]{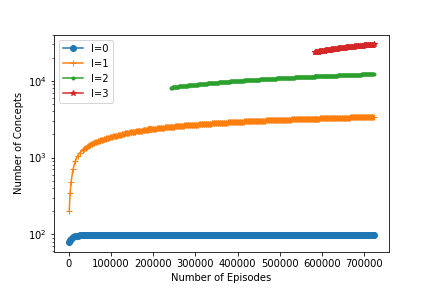} }}
  \subfloat[Number of non-clone concepts with frequency $\ge$ 100.  ]
           {{\includegraphics[height=6cm,width=8cm]{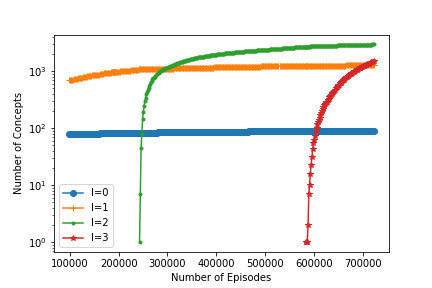}
           }}%
\end{center}
\caption{Number of unique concepts observed (in a segmentation) for
  each level, $l=0, 1, \cdots$), \vs time (number of episodes, or
  lines read), during training of Model4, up to episode 700k. }
\label{fig:cons} %
\end{figure}

Figure \ref{fig:cons} shows the number of (candidate) concepts
generated at each level, as well as the number of concepts seen for at
least 100 times (in a final selected segmentation), \vs the number of
episodes. This is during the training that led to Model4, and the
plots for Model3 are similar.  The number of primitives (unique single
characters) quickly reaches 94, and eventually to 98, in this
abstracts dataset (Fig. \ref{fig:cons}(a)).  The numbers of non-clone
concepts with freq. above 50 for Model4, for levels 0 though 4, were
96, 2.9k, 5.5k, 19.5k and 14k resp., and for Model3 they were
resp. 96, 873, 1.9k, and 12k (up to level 3). Thus, we see a big jump
(10x or more) from level 0 to level 1, and the increase in number of
concepts from one level to next tapers (although non-uniform, and
remains above 3x) for subsequent levels. Recall that for Model3,
concept (composition) generation is on only for the top layer (stopped
at all lower levels). With each additional layer, it appears that we need roughly 10x
more episodes to train the system, under a certain notion of
convergence (see Section \ref{sec:exp_segscores}).

\co{

# the number of concepts that we reached, after so much training..
  
#  eps=2446478 (Model4, laptop1, run2) (run1 ended in 1.2mil)
#  the increases from lev0 to 1, 1 to 2,
# etc. are: ~29x, 2x, 3.5x (for min freq 50) (it does depend when we begin
# the new layer), ...

# level=0 and num non-clone concepts > freq thesh 50 is: 96
# level=1 and num non-clone concepts > freq thesh 50 is: 2864
# level=2 and num non-clone concepts > freq thesh 50 is: 5542
# level=3 and num non-clone concepts > freq thesh 50 is: 19523
# level=4 and num non-clone concepts > freq thesh 50 is: 14237

# level=0 and num non-clone concepts > freq thesh 100 is: 96
# level=1 and num non-clone concepts > freq thesh 100 is: 1478
# level=2 and num non-clone concepts > freq thesh 100 is: 3962
# level=3 and num non-clone concepts > freq thesh 100 is: 12965
# level=4 and num non-clone concepts > freq thesh 100 is: 6939

# eps=1975198 (Model3, laptop2).. the increases from lev 0 to 1, 1 to
# 2, etc. are: ~10x, 3x, 7x (for min freq 50) (it does depend on when
# we begin the new layer), ...

# level=0 and num non-clone concepts > freq thesh 50 is: 96
# level=1 and num non-clone concepts > freq thesh 50 is: 873
# level=2 and num non-clone concepts > freq thesh 50 is: 1892
# level=3 and num non-clone concepts > freq thesh 50 is: 12221

# level=0 and num non-clone concepts > freq thesh 100 is: 96
# level=1 and num non-clone concepts > freq thesh 100 is: 645
# level=2 and num non-clone concepts > freq thesh 100 is: 1824
# level=3 and num non-clone concepts > freq thesh 100 is: 10829

}

\begin{figure}[!htbp]
\begin{center}
  \centering
  \subfloat[ Minimum frequency in an episode.]{{\includegraphics[height=6cm,width=8cm]{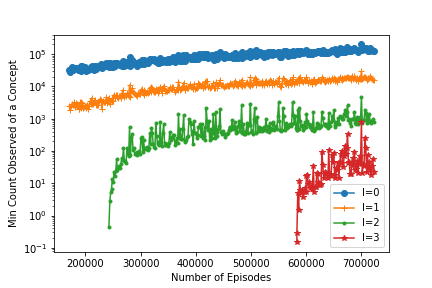}
  }} \subfloat[Further smoothed ]
           {{\includegraphics[height=6cm,width=8cm]{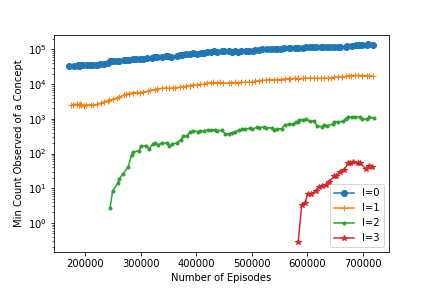}
           }}%
\end{center}
\caption{Minimum of frequency of concepts observed in segmentations
  (moving averages, Model4).}
\label{fig:freqs} %
\end{figure}

\begin{figure}[!htbp]
\begin{center}
  \centering
  \subfloat[For level 2.]{{\includegraphics[height=6cm,width=8cm]{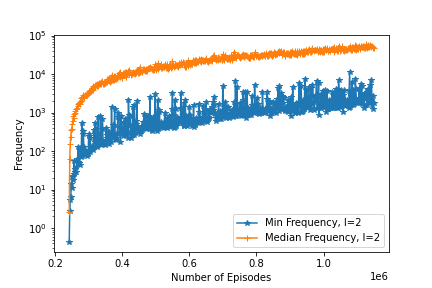} }}
  \subfloat[For level 3.]
           {{\includegraphics[height=6cm,width=8cm]{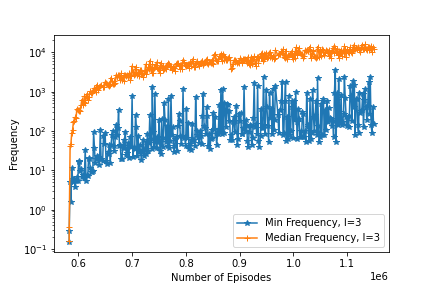} }}
\end{center}
\caption{Median and minimum of frequency of concepts observed in segmentations
  (moving averages, Model4)}
\label{fig:med_freqs} %
\end{figure}

Figure \ref{fig:freqs} shows the moving averages (mixing rate of 0.01)
of the minimum frequency of a concept observed in the final selected
segmentation chain for each level, \vs number of episodes. We see much
variance in the moving average of the minimum
frequency. Fig. \ref{fig:freqs}(b) shows the same plots, but smoothed:
for every 3 consecutive points, the median is kept, and the resulting
is average of the 5 past kept numbers. Fig.~\ref{fig:med_freqs} shows
the moving averages of the median frequencies as well as the minimum
frequenceies, for levels 2 and 3.  We observe that the median is much
higher, perhaps up to two orders of magnitude (100x).

\subsection{Segmentation Scores}
\label{sec:exp_segscores}

\begin{figure}[!htbp]
\begin{center}
  \centering
  \subfloat[Segmentation scores (moving average of \coma)
    ..]{{\includegraphics[height=7cm,width=8cm]
      {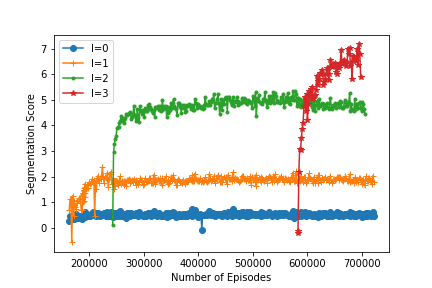} }} \subfloat[Further smoothed: for
    every 3 points, remove high and low y values (keep the middle
    value) and average the last 5 points.]
           {{\includegraphics[height=7cm,width=8cm]{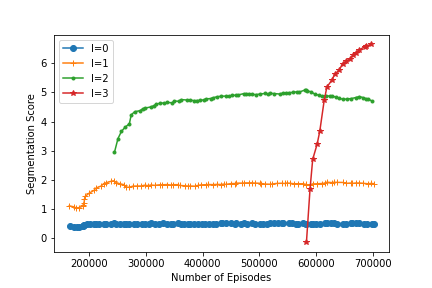}
           }}%
\end{center}
\caption{The progress of segmentation scores (\comap, Eq. \ref{eq:fast}) for various levels ($l=0,
  1, \cdots$) (Model4).}
\label{fig:segscores} %
\end{figure}

Fig. \ref{fig:segscores} shows the progress of segmentation scores at
each level.  As mentioned earlier, we manually trigerred creation of a
new level and segmentation at a new layer, \eg after episode around
600k, we incremented the layers to 3 (\ie allowed segmenting up to
layer $l_{max}=3$).

Note that for these scores, the highest scoring segmentation, out of
several candidates, is picked at the highest level, and the
segmentations at the lower levels, \ie the segmentation chain, are
picked based on which led to the highest segmentation at top.

\co{For
these scores, the actual leave-one-out probabilities received by each
concept in the segmentation is used. However, for picking the highest
scoring segmentation, for the relatively new concepts (occurrence
count below t=50), an optimistic probability of 1 is assumed for them.
}

\begin{figure}[!htbp]
  \hspace*{-0.8cm}
  \subfloat[Level 1. ]
           {{\includegraphics[height=7cm,width=6cm]  {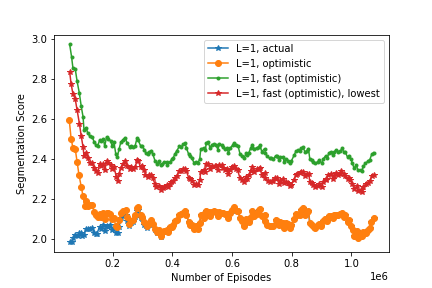} }}
           \subfloat[       Level 3.  ]
                    {{\includegraphics[height=7cm,width=6cm]  {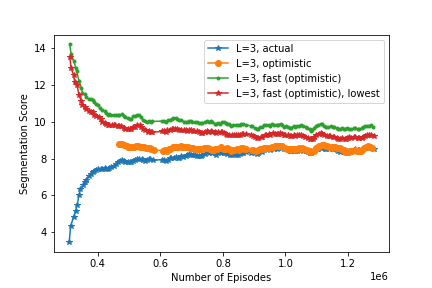} }}
           \subfloat[       Level 4.  ]
                    {{\includegraphics[height=7cm,width=6Cm]  {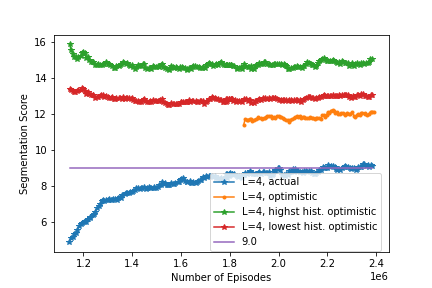} }}
\caption{Optimistic (Eq. \ref{eq:coma2}) \vs actual
  (Eq. \ref{eq:coma}) \vs max \& min historical segmentation scores
  (Eq. \ref{eq:fast}) at the top layer during training that led to
  Model4. Optimistic (for exploration) and actual segmentation scores
  eventually converge, actual segmentation score improving as the
  probability estimates (predictions) improve over time.  For fast
  scoring of candidate segmentations, we use the historical (average
  segmentataion) score of each concept in the segmentation, and those
  values (moving averages) for the picked segmentation (highest
  scoring segmentation in the top layer) as well the lowest scoring
  segmentation in the top-layer (to get a picture of the spread in
  scores) are shown.  }
\label{fig:segscores_vs_opt} %
\end{figure}

We later added reporting of both actual (non-optimistic) and
optimistic segmentation (\comap) scores.  See
Fig. \ref{fig:segscores_vs_opt}. Optimistic \coma starts higher,
but eventually once most concepts are seen frequently enough (some
time after $c.freq > t_{opt}$ for most observed concepts $c$), the
optimistic and actual (non-optimistic) \coma scores converge.  For
Model4 (trained up to level 4), we note that the scores at the top, level 3,
have not converged yet, after 2 million episodes. Our experience on
this dataset, and with our current parameter settings (such as
$t_{opt}=50$), suggests that for the actual and optimistic
segmentation scores to converge to say 10\% to 20\% of one another
each level requires roughly 10x more episodes than the level below for
convergence (an order of magnitude larger). Thus level $0$ requires
about 10k to 20k episodes (see also Table \ref{tab:rate_coma} in the
appendix), level 1 requires 100k to 200k, level 2 about 1 million to 2
million, and we expect level 3 requires about 10million episodes.


\subsection{Quadratic Loss and Number of Concepts Per Episode}
\label{sec:exps_qloss}

\begin{figure}[!htbp]
\begin{center}
  \centering
  \subfloat[Squared loss on probabilities]{{\includegraphics[height=8cm,width=8cm]
      {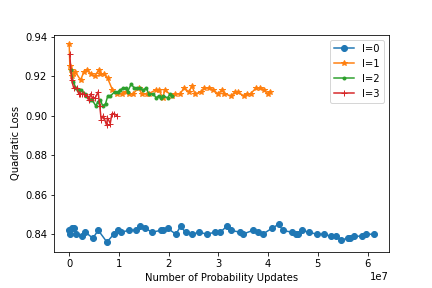} }} \subfloat[Number of concepts per episode, at each level.  ]
           {{\includegraphics[height=8cm,width=8cm]{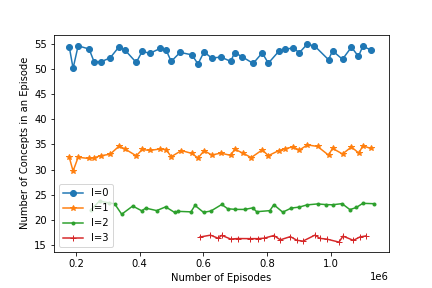}
           }}%
\end{center}
\caption{Progress in quadratic loss and average number of concepts in
  in each level of the selected chain per episode, over time. The
  number of primitives (level 0 concepts) is about 54 per episode. The
  convergence is fast and these measures are only crude ways of
  measuring system progress, compared to the \coma score over time
  and for various levels (Figs. \ref{fig:segscores_vs_opt} and
  \ref{fig:segscores}). }
\label{fig:qloss} %
\end{figure}

Fig. \ref{fig:qloss}(a) shows progress in the moving average over
episodes of quadratic loss at each level. Once a segmentation chain is
picked, and probabilities are computed in take-each-out manner (see
GetProb() in Fig. \ref{fig:scoring}), we can compute the loss for each
position and average (for each level separately).  We are reporting
the loss, $(1.0-p)^2$, on the probability $p$ assigned to the true
target concept (at each level). $p$ can be as low as 0 here, thus the
maximum loss can be 1.0.  We see that convergence is fast, and the
loss goes to around 0.84 for level 0 and to 0.9 loss for other higher
levels. Quadratic loss is only a crude way of measuring progress of
the system.

Fig. \ref{fig:qloss}(b) shows the moving average over episodes of the
average number of concepts in the selected segmentation chain. There
are about 53 primitive concepts (raw characters in a line of an
abstract) on average. Again, we see convergence is fairly fast (while
\coma keeps steadily improving).

\subsection{Example Holonyms}
\label{sec:exps_holo}

Table \ref{tab:holos1}, for Model3, shows the number of holonyms and
some example holonyms, at different levels, ranked by their frequency
for a few concepts, corresponding to the frequent "a" and infrequent
"z" as well as "there".  Recall that for Model3, co-occur statistics
are only updated and used at the top layer, and holonyms can only be
added at the top layer.

\begin{table}[!htbp]\center
  \begin{tabular}{ |r|r|r|r|r| }     \hline
     $\con_0($'a'$)$ (31) & $\con_1($'a'$)$ (86) & $\con_2($'a'$)$ (308) & $\con_0($'z'$)$ (6) & $\con_2($'ther'$)$(62) \\ \hline
 1. 805558 "al" &  1. 442829 "and" &  1. 135721 "an" &  1. 51935 "ze" & 1. 10932 "other" \\ 
 2. 684580 "ar" &  2. 146546 "ati" &  2. 123130 "at" & 2. 34126 "za" & 2. 4024 "there" \\ 
 3. 594014 "ra" &  3. 120726 "gra" &  3. 75276 "ca" & 3. 11007 "zi" & 3. 3383 "whether" \\ 
 4. 397963 "ma" &  4. 104855 "da" & 4. 56968 "pa" & 4. 4385 "zy" & 4. 2435 "gether" \\ 
 5. 290619 "la" &  5. 88762 "ica" &  5. 36479 "ation" & 5. 947 "Hz" & 5. 2083 "thern" \\ 
 6. 241694 "ha" &  6. 78547 "cha" &  6. 18858 "orta" & 6. 492 "zz" & 6. 1908 "further" \\
 \cline{4-4}  
 7. 156664 ",a" &  7. 72637 "sta" &  7. 15369 "sa" & $\con_1($'z'$)$ (4) & 7. 1752 "thers" \\ \cline{4-4}  
 8. 96328 "ga" &  8. 67244 "fa" &  8. 14550 "gra" & 1. 4969 "zon" & 8. 1688 "thermo" \\ 
  20. 9997 "ca" &  9. 51748 "rea" &  24. 5323 "Ca" & 2. 4969 "zo" & 16. 597 "mother" \\ 
 21. 9061 ";a" &  43. 1853 "Ea" &  25. 5171 "colla" & 3. 380 "zer" & 17. 518 "therthe" \\ 
 22. 5969 "Ba" &  44. 1818 "Ala" &  26. 5090 "media" & 4. 0 "riz" & 18. 515 "gather" \\ 
\cline{4-4} 24. 4583 "pa" &  45. 1669 "gya" &  245. 51 ".Oura" &  $\con_2($'z'$)$ (12) & 28. 308 "Another" \\ \cline{4-4}
 28. 2209 "Va" &  46. 1572 "ja" &  246. 50 "Gia" & 1. 218 "z," & 38. 168 "Norther" \\ 
 29. 632 "Ca"  &  84. 7 ",ma" &  300. 2 "bona" & 2. 172 "artz" & 39. 167 "ngother"\\ 
 30. 387 "Pa" & 85. 3 "yba" &  307. 0 "alpra" & 3. 146 "zma" & 59. 4 "therfie"  \\
 31. 50 "Ya" & 86. 2 "ywa" &  308. 0 ".Ana" & 10. 0 "zone" & 62. 0 "theras" \\ \hline
  \end{tabular}
\vspace{.2cm}
\caption{Examples of the holonyms of a few concepts from Model3 (in
  the next level), sorted by their frequency.  Each holonym's
  frequency and its rank order is also shown. The concept corresponding
  to "a" at level 0 has 31 holonyms in Model3 (at level 1), and its
  most frequent holonym is "al" ($\con_1($'al'$)$) with frequency 805k. }
\label{tab:holos1}
\end{table}

For Model4, recall that holonyms could be added into any layer, but
with the constraint that we did not allow the creating parts to be
both clones. We note that the number of holonyms of a concept (at the
next layer) can be significantly higher, which can be an issue for
efficiency.  In the implementation, we limited the list of
co-occurrences, but not the number of holonyms of a concept. If a
concept is frequent, such as those corresponding to "a", the less
frequent concept to the left can make compositions with it, and we see
this for $\con_2($'a'$)$ and $\con_3($'a'$)$ in Table \ref{tab:holos2} for
Model4 (\ie many holonyms are a concept followed by "a", such as
"understa", "thatca", and "pathwa"). This appears less so in Table
\ref{tab:holos1} for Model3, where we allowed cloned concepts to
join. We also note that the number of holonyms of $\con_1($'a'$)$ and
$\con_2($'a'$)$ in Model3 appear lower than in Model4 (nearly half or
less): part of the reason is Model4 was trained for several 100k
episodes more, but another reason is likely turning off compositions
at lower layers for Model3.

In Table \ref{tab:holos1}, also observe that many holonyms at higher
layers, such as $\con_3($'ca'$)$ had been discovered before, but they
are created again. Looking into it, $\con_2($'ca'$)$ and $\con_1($'ca'$)$,
have relatively low scores, and in particular lower than the parts
(the clones) $\con_2($'a'$)$ and $\con_2($'c'$)$, and thus are likely not
picked during segmentation at their layers. So they are created again
(rediscovered). We also note that we currenly do not connect clones to
their (cloned) parts at higher layers (thus $\con_2($'ca'$)$, the clone
of $\con_1($'ca'$)$, does not connect to $\con_1($'a'$)$ and
$\con_1($'c'$)$). Thus, the system, with our current implementation, may
need to redicover some holonyms at higher layers (\eg $\con_3($'ca'$)$),
specially if a holonym is not picked at a lower layer due to low
scores (as is the case with $\con_2($'ca'$)$).

\begin{table}[!htbp]\center
  \begin{tabular}{ |c|lc|lc|lc|lc| }     \hline
$con_3($'a'$)$ (1069) & 1. 1942 & understa & 2. 1368& thatca & 3. 1351&  arthqua & 4. 1137 & pathwa \\ \hline
$con_2($'a'$)$ (594) & 1. 17467 & orga & 2. 10679& area  & 3. 9701 & mecha & 4. 6446 & inga \\ \hline
$con_1($'a'$)$ (332) & 1. 492520 & and & 2. 83939& are  & 3. 82188 & cha & 4. 74683 & ant \\ \hline
$con_0($'a'$)$ (92) & 1. 1468312 & at & 2. 994309 & al  & 3. 700035 & ar & 4. 463557 & ra \\ \hline
  \end{tabular}
\vspace{.2cm}
\caption{Top holonyms of "a" in different levels of Model4, with
  similar information to Table \ref{tab:holos1}, but shown left to
  right. $con_3($'a'$)$ has 1069 holonyms in Model4, and its most
  observed holonym is "undresta" ($con_4($'undresta'$)$) with
  freq. 1942.}
\label{tab:holos2}
\end{table}

\co{

Model 4:

concept is: "a" lev=3

# num holonyms: 1069
# 1. 1942 understa
# 2. 1368 thatca
# 3. 1351 arthqua
# 4. 1137 pathwa
# 5. 1107 ationa
# 6. 873 imulta
# 7. 873 approa
# 8. 844 phenomena
# 9. 795 theba
# 10. 644 thatma
# 11. 623 edtoa
...
# 349. 52 asesa
# 350. 52 ariseina
# 351. 51 verna
# 352. 51 stoobta
...
# 1040. 1 -of-thea
# 1041. 0 wideba
# 1042. 0 udes:a

concept is: "a" lev=2

# num holonyms: 594
# 1. 17467 orga
# 2. 10679 area
# 3. 9701 mecha
# 4. 6446 inga
# 5. 5876 data
# 6. 5767 beca
# 7. 5018 obta
# 8. 4813 funda
...
# 331. 79 usua
# 332. 78 sexpa
# 333. 78 loya
# 334. 76 Infra
# 335. 76 'sla
...
# 568. 1 leofa
# 569. 1 dtoba
# 570. 1 berea
# 571. 1 Loga
# 572. 1 -bea
# 573. 0 tymea
# 574. 0 thena
...

concept is: "a" lev=1

# num holonyms: 332
# 1. 492520 and
# 2. 83939 are
# 3. 82188 cha
# 4. 74683 ant
# 5. 56778 rea
# 6. 55256 ach
# 7. 51831 sta
# 8. 48053 ase
# 9. 46502 pla
# 10. 46462 act
# 11. 42697 tra
# 12. 30116 nda
# 13. 29988 rta
# 14. 26302 gra
      ..
      # 207. 1 Bla
# 208. 1 )ma
# 209. 1 )Fa
# 210. 0 zea
# 211. 0 ywa
# 212. 0 yva

concept is: "a" lev=0

# num holonyms: 99
# 1. 1468312 at
# 2. 994309 al
# 3. 700035 ar
# 4. 463557 ra
# 5. 203792 ab
# 6. 187135 pa
...
# 96. 50 0a
# 97. 50 
# 98. 37 ]a
# 99. 17 *a

}
    
\subsection{Examples and Some Statistics of the Prediction Edges}
\label{sec:exps_ex}

Top 10 prediction edges for six concepts in Model3, for positions -1,
1, and 2, in levels 0 or 1, and level 3, are shown in Table
\ref{tab:top10}. As explained in Section \ref{sec:edges} and Appendix
\ref{app:dlr} the edge weights approximate conditional probabilities
and for a given position (such as position 1), they sum to at most
1.0.  Many of the higher edge weights suggest (parts of) common words
and phrases being formed, \eg immediately after "th" in layer 1, "er"
occurs with high probability, which, once composed, can eventually
lead to say "there" in a higher level. Note also how the distributions
change for the same pattern at the different levels.  At level 3, "er"
is no longer in top 10, probably because "ther" has been created
already, and the segmenter uses it in segmentation.

Table \ref{tab:concepts} shows that "ther" is indeed created in level
2. The table also shows a few related concepts, "there" and "whether",
as well as the five highest frequency (most seen) non-clone concepts
and the most frequent clone concept, $\con_3($'s'$)$ in level 3. The
most seen concept in level 3 is $\con_3($'sand'$)$. The word "sand" is
not common in the corpus (occurs 100s of times only, but seen 54k times by
Model3), and from looking at the top prediction weights of
$\con_3($'sand'$)$, for instance "tie", "method", "alysi" appear
before $\con_3($'sand'$)$, we conclude that likely this is an example
of putting together two concepts, "s" (possibly a pluralization of
previous nouns) and "and". Sect. \ref{sec:exps_segs} looks the count
and ratio of bad splits (\eg the term "calls" being split into "call"
and "s") and we observe that such error statistics do improve with
more learning and inference.  However, it is an open question whether
these errors go to zero with more learning (does the system recover
from {\em all} past mistakes?)  and/or whether we need better
algorithms (learning and inference). Sect. \ref{sec:exps_segs}
discusses the effect of the search width as well.

In Model 3, the top predictions of the begin-buffer predictor
(Sect. \ref{sec:special}) in level 3 were, in order, "e", "a", "s",
"in", "the", "Thi", "and", and "The" (from weight of 0.041 to 0.015).,
while at level 2, they were, "a", "i", "e", "t", "s", "the", "pro",
"The" (from weight of 0.049 to 0.026).

\begin{table}[htbp]\center
  \begin{tabular}{ |c|c|c|| c|c|c| }     \hline
      pos=-1 & pos=1 & pos=2 & pos=-1 & pos=1 & pos=2 \\ \hline
     \multicolumn{3}{ |l||}{\hspace*{1.3cm} $con_0($'a'$)$ ('a' at level 0)} &
    \multicolumn{3}{l|}{\hspace*{1.5cm} $con_3($'a'$)$ ('a' at level 3)}  \\ \hline
  "r",0.107  &  "n",0.200  &  "i",0.155 & "s",0.076  & "s",0.091  & "a",0.034 \\
 "e",0.101  &  "t",0.191  &  "e",0.146 & "t",0.063  & "te",0.025  & "s",0.034  \\
 "t",0.079  &  "l",0.143  &  "d",0.125 & "n",0.059  & "in",0.018  & "e",0.025 \\ 
 "n",0.074  &  "r",0.120  &  "t",0.087 & "dat",0.038  & "i",0.018  & "t",0.021 \\ 
 "c",0.073  &  "s",0.065  &  "a",0.064 & "w",0.037  & "c",0.017  & "n",0.019 \\ 
     "h",0.070  &  "c",0.059  &  "o",0.050 & "e",0.029  & "d",0.016  & "o",0.015 \\ 
 "l",0.070  &  "b",0.029  &  "c",0.048 & "te",0.022  & "ble",0.014  & "i",0.012 \\ 
 "m",0.065  &  "d",0.027  &  "s",0.047 & "an",0.017  & "nt",0.013  & "to",0.012 \\ 
 "s",0.057  &  "m",0.027  &  "l",0.044 & "to",0.016  & "re",0.011  & "the",0.011 \\ 
 "p",0.034  &  "p",0.025  &  "n",0.026  & "in",0.016  & "ch",0.009  & "d",0.010 \\ \hline
     \multicolumn{3}{ |l||}{\hspace*{1.3cm} $con_1($'th'$)$  } &
    \multicolumn{3}{l|}{\hspace*{1.5cm} $con_3($'th'$)$ }  \\ \hline
  "wi",0.215  & "er",0.146  & "t",0.103 & "e",0.066  & "at",0.287  & "the",0.039 \\ 
  "o",0.072  & "a",0.143  & "u",0.073  & "o",0.053  & "is",0.052  & "a",0.038 \\ 
  "me",0.067  & "o",0.125  & "s",0.070  & "dep",0.043  & "s",0.047  & "s",0.034 \\ 
  "bo",0.039  & "e",0.122  & "d",0.069  & "Sou",0.038  & "us",0.037  & "e",0.024 \\ 
  "ng",0.034  & "t",0.053  & "e",0.059  & "a",0.037  & "an",0.035  & "t",0.019 \\ 
  "a",0.032  & "i",0.049  & "n",0.049  & "pa",0.034  & "n",0.028  & "ca",0.014 \\ 
  "i",0.032  & "ro",0.046  & "he",0.048  & "veleng",0.021  & "erin",0.018  & "f",0.013 \\ 
  "e",0.029  & "re",0.037  & "a",0.042  & "algori",0.018  & "no",0.018  & "n",0.013 \\ 
  "or",0.027  & "m",0.023  & "i",0.031  & "workwi",0.018  & "r",0.017  & "ust",0.013 \\ 
  "d",0.027  & "s",0.021  & "se",0.027  & "leng",0.017  & "in",0.015  & "ic",0.012 \\ \hline
     \multicolumn{3}{ |l||}{\hspace*{1. cm} $con_2($'ther'$)$ }  &
     \multicolumn{3}{l|}{\hspace*{1. cm} $con_3($'ther'$)$ }  \\ \hline
   "o",0.311  & "e",0.100  & "a",0.050 & "i",0.168  & "mody",0.058  & "n",0.076 \\ 
  "whe",0.076  & "s",0.067  & "e",0.038  & "wea",0.096  & "es",0.028  & "e",0.042 \\ 
  "ge",0.063  & "mo",0.050  & "s",0.037  & "ando",0.086  & "mos",0.022  & "a",0.034 \\ 
  "fur",0.050  & "n",0.048  & "n",0.035  & ",ando",0.066  & "p",0.019  & "s",0.025 \\ 
 "u",0.036  & "t",0.042  & "t",0.032  & "so",0.041  & "a",0.018  & "t",0.025 \\ 
  "i",0.033  & "the",0.036  & "o",0.026  & "Wea",0.026  & "mal",0.018  & "o",0.020 \\ 
  "a",0.030  & "mal",0.029  & "i",0.021  & "e",0.022  & "b",0.017  & "i",0.015 \\ 
  ".Fur",0.026  & "i",0.027  & "se",0.019  & "O",0.020  & "pre",0.013  & "y",0.014 \\ 
  "ro",0.025  & "th",0.024  & "dy",0.019  & "ur",0.019  & "lands",0.012  & "p",0.013 \\ 
  "ra",0.025  & "more",0.021  & "g",0.014  & "sei",0.019  & "to",0.012  & "te",0.011 \\ \hline
  \end{tabular}
\vspace{.2cm}
  \caption{Top 10 edge weights for a few concepts and positions: the
    concepts, $con_0('a')$, are placed in the table so the -1 edges
    are slighlty to the left, and +1 to the right, for each of reading
    the implied relations.  For $con_0('a')$, \ie 'a' at level 0, we
    get 'r' ($con_0('r')$) occurring before it with highest
    probability (0.1) and 'n' ($con_0('n')$) immediately after it. For
    $con_1('th')$, we get "wi" in level 1 at position -1 with high
    weight, or together they would be "with", and "er" at position 1
    with high weight ("ther"). At level 3, we observe "dep" ("depth")
    and "workwi" among the high weights.  }
\label{tab:top10}
\end{table}

\co{
\begin{table}[!htbp]\center
  \begin{tabular}{ |c||c|c|c|| c|c|c| }     \hline
     & pos=-1 & pos=1 & pos=2 & pos=-1 & pos=1 & pos=2 \\ \hline
    & \multicolumn{3}{ c||}{$con_2($'ther'$)$ level=2}
    & \multicolumn{3}{l|}{$con_3($'ther'$)$ level=3}  \\ \hline
  & "o",0.311  & "e",0.100  & "a",0.050 & "i",0.168  & "mody",0.058  & "n",0.076 \\ 
 & "whe",0.076  & "s",0.067  & "e",0.038  & "wea",0.096  & "es",0.028  & "e",0.042 \\ 
 & "ge",0.063  & "mo",0.050  & "s",0.037  & "ando",0.086  & "mos",0.022  & "a",0.034 \\ 
 & "fur",0.050  & "n",0.048  & "n",0.035  & ",ando",0.066  & "p",0.019  & "s",0.025 \\ 
"ther" & "u",0.036  & "t",0.042  & "t",0.032  & "so",0.041  & "a",0.018  & "t",0.025 \\ 
 & "i",0.033  & "the",0.036  & "o",0.026  & "Wea",0.026  & "mal",0.018  & "o",0.020 \\ 
 & "a",0.030  & "mal",0.029  & "i",0.021  & "e",0.022  & "b",0.017  & "i",0.015 \\ 
 & ".Fur",0.026  & "i",0.027  & "se",0.019  & "O",0.020  & "pre",0.013  & "y",0.014 \\ 
 & "ro",0.025  & "th",0.024  & "dy",0.019  & "ur",0.019  & "lands",0.012  & "p",0.013 \\ 
 & "ra",0.025  & "more",0.021  & "g",0.014  & "sei",0.019  & "to",0.012  & "te",0.011 \\ \hline
  \end{tabular}
\vspace{.2cm}
  \caption{Top 10 edge weights for positions -1, 1, 2, and levels 2 and 3, for "ther".}
\label{tab:top10_2}
\end{table}
}

\begin{table}[!htbp]\center
  \begin{tabular}{ |c|c|c|c||c|c|c|c| }     \hline
    concept    & freq.  & last seen & score$^*$
    & concept & freq.  & last seen & score \\ \hline
 $con_2($'ther'$)$  & 57890 & 25 & 14.0 & $con_3($'sand'$)$ & 54456 & 24 & 10.1 \\ \hline
 $con_3($'ther'$)$  & 6370 & 106 &  8.0 & $con_3($'research'$)$ & 50353 & 20 & 25.3 \\ \hline
 $con_3($'there$')$ & 4023 & 58 &  10.4 & $con_3($'project'$)$ & 42501 & 22 & 28.0 \\ \hline
 $con_2($'with'$)$  & 84643 & 16 &  17.1 & $con_3($'ation'$)$ & 36479 & 101 & 13.4\\ \hline
 $con_3($'with'$)$  & 22195 & 48 &  10.8 & $con_3($'develop'$)$ & 28092 & 85 & 25.5 \\ \hline
  $con_3('whether')$  & 3383 & 388 &  21.0 & $con_3($'s'$)$ & 966729 & 2 & 1.1 \\ \hline
 \end{tabular}
\vspace{.2cm}
\caption{Statistics on a few concepts in Model3: left columns show
  concepts ("ther") or related ones ("whether") from Table
  \ref{tab:top10} and right columns show a few most frequent. The
  frequency (number of times seen), how many episodes ago it was last
  seen in a segmentation, from the time the snapshot was taken (\eg
  $con_2('ther')$ was seen 22 episodes ago), and the average
  historical reward are shown. The right column shows non-clone
  concepts with highest freq.  at level 3, except the last row,
  $con_3('s')$, which is the concept with the highest freq. in level 3
  (it is a clone).  }
\label{tab:concepts}
\end{table}

We next explore a few properties of the prediction edges, in
particular properties of the probabilities and measures of uncertainty
such as average entropies.  Table \ref{tab:entr} shows average
entropies (averaged over concepts)\footnote{Entropy is $-\sum_p
  p\log(p)$, or $ \sum_{c_j} w_{c_i,c_j,k}\log(w_{c_i,c_j,k})$ (where
  the sum goes over the edges of a concept $c_i$ for a position $k$),
  where, where for any position $k\in \Delta$, $\sum_{c_j} w_{c_i,c_j,k} \le 1$
  (See Appendix \ref{app:dlr}).  We return 0 entropy when there are no
  edges, \eg for concepts generated but not seen yet in a
  segmentation. } for concepts that have been observed for some time
(under three different freq. thresholds), for Model3.  Position 1 (pos=1) is
the next (immediate right side) concept. In general, entropy increases
as the concept is seen more, \ie the diversity of concepts it is seen
with grows. Note that our rate decay schedule based on count of
updates also leads to lower entropy (larger edge weights or
probabilities) for the less-seen concepts.


\begin{table}[!htbp]\center
  \begin{tabular}{|c| c|c|c|c|| c|c|c|c|| c|c|c|c|}     \hline
    level &  N & pos=1 & pos=2 & pos=3 & N, 50 & pos=1 & pos=2 & pos=3 & N, 100 & pos=1 & pos=2 & pos=3   \\ \hline
    3 & 20.4k & 2.26 & 3.09 & 3.35 & 14.8k & 2.84 & 3.91  & 4.26 & 12.9k & 2.96 & 4.06 & 4.42  \\ \hline
    2 & 3.85k & 2.27 & 3.02 & 3.34  & 2.6k &  2.82 & 3.84 & 4.28 & 2.5k & 2.84 & 3.87 & 4.32  \\ \hline
    1 & 973 & 3.1 & 3.57 & 3.74 & 969 & 3.12 & 3.58 & 3.75 & 741 & 3.19 & 3.73 & 3.92 \\ \hline
    0 & 98 & 2.66 & 2.95 & 3.05  & 96 & 2.70 & 3.0 & 3.1 & 96 & 2.70 & 3.0 & 3.1 \\ \hline
\end{tabular}
\vspace{.2cm}
  \caption{Entropy of edge weights (conditional probabilities) in Model3,
    averaged over concepts of a level, where N is the number of
    concepts that remain with $\ge$ 0, 50 and 100 thresholds on
    frequency. Thus, for this model at level 3 there are 20.4k
    concepts generated, while 14.8k concepts remain if we require a
    concept to have been seen (in a final selected segmentation) at
    least 50 times. Entropy goes up with position (the more distant,
    the more uncertain), and often also with level (the more concepts,
    the more uncertain), and with higher concept freq. (experience
    with the concept), \ie more concepts are seen with more
    experience. }
\label{tab:entr}
\end{table}

The average entropies for negative positions are similar, \eg pos=-1
is similar to pos=1 (\eg average entropy of 2.90 for pos=-1 for freq
$\ge$ 50, and 3.1 for freq. $\ge$ 100, at level 3). Medians are
similar to the means. Minimum entropies can be 0 (those concepts that occur
with just one other concept in the corpus).


Table \ref{tab:mass} shows a similar pattern with respect to
probability mass. The table gives the sum of probabilities on edges,
for those edges above the probability threshold (shown for $\ge$ 0.01
and 0.1), \ie how much of the mass is concentrated on the probability
ranges we expect are most useful (ratios to total probability mass are
similar, a bit larger).  Higher levels in general have
higher entropy and lower mass on high probability edges, and position
2 is more uncertain than pos 1.

Table \ref{tab:mass} also presents the average number of edges for
several cases. We note that the number of edges with relatively high
weight (above 0.01) are in the 10s, and we expect this has
implications with regards to efficiency of prediction and inference,
\ie the numbers of significant edges are relatively small, which can
speed up operations (specially whenever learning is turned off). Also,
we note that as the average edge numbers do not seem to be increasing
with higher levels, while the number of concepts substantially
increase with level, the sparsity of connections, the graph sparsity,
increases with level.  The average number of edges over all concepts
(under no freq. threshold on concepts) is not shown, but those
averages, under any weight threshold, are smaller than the numbers
shown in the table for concepts passing the frequency thresholds.

\begin{table}[htbp]\center
  \begin{tabular}{|c|c|c|c|c|c|c|c|c|c|c|c|  }     \hline
    & $p\ge 0.0$  & \multicolumn{2}{c|}{$p\ge 0.01$}
    &  $p\ge 0.0$  & \multicolumn{2}{c}{$p\ge 0.01$}
    & \multicolumn{4}{|c|}{$p\ge 0.10$}  \\ \hline
    &  \multicolumn{3}{c}{freq. $\ge 50$} & \multicolumn{3}{|c|}{freq. $\ge 100$} &
    \multicolumn{2}{|c|}{freq. $\ge 50$} & \multicolumn{2}{|c|}{ freq. $\ge 100$}  \\ \hline
    level & pos=1 & pos=1 & pos=2 & pos=1 &  pos=1 & pos=2  & pos=1 & pos=2 & pos=1 & pos=2 \\ \hline
    3 & 87  & 0.80, 16.6  & 0.64 & 97 & 0.78, 16.5 & 0.59 & 0.41, 1.46  & 0.15 & 0.39, 1.4 & 0.13  \\ \hline
    2 & 129  & 0.81, 15.2 & 0.65 & 134 & 0.80, 15.0  & 0.63 & 0.44, 1.6 &  0.2 & 0.44, 1.6 & 0.2 \\ \hline
    1 & 112 & 0.84, 20.6 & 0.78 & 139 & 0.79, 18.9 & 0.72 & 0.31, 1.44 & 0.16 & 0.32, 1.46 & 0.16   \\ \hline
    0 & 70 & 0.90, 17.2 & 0.89 & 70 & 0.90, 17.3 & 0.89 & 0.44, 2.3 & 0.28 & 0.44, 2.3  & 0.28  \\ \hline
\end{tabular}
\vspace{.2cm}
  \caption{Average (over concepts) of edge-weight (probability) mass
    over prediction edges with weight surpassing 0.01 and 0.1
    thersholds for pos=1 and pos=2.  The average number of such edges
    is also shown for pos=1, and thresholds 0 (no threshold), 0.01 and
    0.1. Thus at level 3, for concepts with freq. $\ge$ 50, on average
    there are 87 edges per concept, and about 17 edges for such
    concepts have a weight of no less than 0.01, and this average is
    1.4 for thresh. 0.1 (all for pos=1). We observe the number of
    edges goes up with growing concept freq. (from 50 to 100) but the
    number of edges with high weight can slightly go down. Across
    layers, the total mass of edges does not change much, and much of
    the mass, around $80\%$ or more, are on edges with
    weight $\ge 0.01$ for pos=1. }
\label{tab:mass}
\end{table}

\subsection{Example Segmentations, and Statistics on Bad splits}
\label{sec:exps_segs}

Table \ref{tab:seg_examples} shows a couple of lines and their
selected segmentations and segmentation chains (down to levels 1 and
2), via Model3 and Model4.  Each concept in a segmentation corresponds
to a consecutive sequence of characters in the input line, and
therefore has a corresponding beginning and an end. As concepts are
created and segmented with, we hope that these boundaries increasingly
correspond to the original blank spaces (separations) in the input line.


Here we count the number of "{\bf bad splits}" on average.  When
looking at the sequence of active concepts in the top level, a split
is the location of an end of one concept and beginning of the next
one, (thus, we don't count the beginning and end of lines).  A {\bf
  \em bad1} split is a split within a word (or that does not align
with the space between two words). For instance in the top level
selected segmentation from Table \ref{tab:seg_examples}, in the prefix
"reg arding the conser", there is 1 bad1 split, between "reg" and
"arding" and 2 good splits (between "arding" and "the", and between
"the" and "conser").  The whole selected segmentation via the level 3
model has 5 bad1 splits.  A {\bf \em bad2} split is a bad split on
both sides, \ie it is a bad1 split where the next split, as we scan
from left to right, is also bad (thus both ends of a concept is
internal to one or two words).  Table \ref{tab:splits} reports on the
number of bad1 and bad2 splits as well as \coma scores, for two models
and several beam width parameters, averaged over about 200 lines. The
table also includes the average segmentation scores, for different
beam widths.  Level 1 model was trained at 160k episodes, Model3 and
Model4 were trained for nearly 2 million episodes each.  We observe
increased \coma correlates with fewer bad splits.  And with more
levels, in general the \coma score goes up and bad splits go down.
And of course, the more search trials, \ie the wider the beam search,
the better the results. Our default of 10 tries, keep 3 yields similar
to the 5,5 row in the Table. With 10,10 and Model4, bad1s and bad2s go
to 6.8 and 3.6 resp., and with 15,15 bad1 and bad2s go to 6.5 and 3.5
resp. with \coma of around 11.5 in both cases. Multiple trials give
similar results.  With additional training and layers, we have
observed number of bad splits to go below 3.  We note that on average
a line has about 8 separating blank spaces ("true" splits) or 9
space-separated tokens.

\co{
\begin{figure}
\begin{verbatim}
Line = "regarding the conservation and management of these magnificent"
level 3: reg arding the conser vation and mana gement ofthese magni fice nt , via a level 3 model
level 2: re g ar ding the con ser va tion and ma na ge ment oft hese mag ni fice nt 
level 1: re g ar di ng t he c on se r va ti on a nd ma n a ge me nt o ft he se ma g n i fi ce nt 
level 4: regardingthe conserv ationa ndman a gementofthe s emagni fic ent , via a level 4 model
level 3: regard ingthe conser v ation a ndma n a gementof the s emagni fic ent 
level 2: reg ard in gthe cons er v atio n a ndm a n a geme ntof the s ema gni fic ent 
\end{verbatim}
\end{figure}
}

\begin{table}[htbp]
  \begin{tabular}{ |c|c|l| }     \hline
\multicolumn{3}{|c|}{ line = "regarding the conservation and management of these magnificent"}\\
\multicolumn{3}{|c|}{(blanks removed: regardingtheconservationandmanagementofthesemagnificent)}\\\hline
    & L=3  & reg arding the conser vation and mana gement ofthese magni fice nt \\
Model3, &    L=2  & re g ar ding the con ser va tion and ma na ge ment oft hese mag ni fice nt \\
11.7 score     &    L=1  &  re g ar di ng t he c on se r va ti on a nd ma n a ge me nt o ft he se ma g n i fi ce nt  \\ \hline
  & L=4  & regardingthe conserv ationa ndman a gementofthe s emagni fic ent \\
Model4, &  L=3  & regard ingthe conser v ation a ndma n a gementof the s emagni fic ent \\ 
17.3 score  & L=2  & reg ard in gthe cons er v atio n a ndm a n a geme ntof the s ema gni fic ent \\ \hline
\multicolumn{3}{|c|}{ line = Commercial exploitation over the past two hundred years drove }\\\hline
& L=3 & Commer cial e xploi t ationo verthe pa st t w o h undred y e ars d rove \\
Model3, & L=2 & Comm er ci al e xplo i t ati ono ver the p a st t w o h und red y e ar s d rove \\ 
6.3 score & L=1 & Co mm er c i al e xp lo i t a ti on o v er t he p a st t w o h u nd re d y e ar s d ro ve \\ \hline
&  L=4 & Commercial e xploitatio n overthe pasttwo hundred y e ar s dro ve \\

Model4, &  L=3 & Comm ercial e xploi tatio n overthe pasttwo hundre d y e ar s dro ve \\

11.9 score & L=2 & Comm er cial e xplo i t atio n ove rthe past two hun dre d y e ar s dro ve \\\hline
  \end{tabular}
  \vspace*{.1in}
  \caption{Segmentation of a couple of example lines (episodes) via
    two models, using width 15 (15 tries from each, and keep highest
    scoring 15 at each level). The segmentation chains covering a few
    lower levels are also shown.}
  \label{tab:seg_examples}
\end{table}

\begin{table}[htbp]
  \begin{tabular}{ |c|l| }     \hline
    Model4 & Commercial(0.017) e(0.236) xploitatio(0.016) n(0.078) overthe(0.038) \\
    score: 11.9 & pasttwo(0.011) hundred(0.000) y(0.076) e(0.148) ar(0.074) s(0.083) dro(0.000) ve(0.002) \\ \hline
    Model3 &  Commer(0.002) cial(0.07) e(0.157) xploi(0.002) t(0.191) ationo(0.004)
    verthe(0.002) pa(0.057) st(0.051) \\
    score: 6.3 & t(0.074) w(0.052) o(0.095) h(0.180) undred(0.001) y(0.145) e(0.206)
    ars(0.040) d(0.004) rove(0.000) \\ \hline
    Model3 & Com(0.037) merci(0.004) ale(0.001) xploi(0.002) t(0.17) ationo(0.004) verthe(0.002) pa(0.057)  st(0.051)\\
    score: 5.9 & t(0.074) w(0.052) o(0.095) h(0.180) undred(0.001) y(0.145) e(0.206) ars(0.040) d(0.004) rove(0.000) \\ \hline
  \end{tabular}
  \vspace*{.1in}
  \caption{Segmentation examples, shown with the assigned
    probabilities the active concepts received (in the take-each-out
    fashion).  The segmentations are of the same line, one by Model4
    and two by Model3 (at the top layer of the model), along with
    score of the segmentation (11.9 by Model4, 6.3 and 5.9 by
    Model3)).  }
  \label{tab:seg_examples2}
\end{table}

\co{
when a concept left or right boundary did not coincide with a blank
space or start or end of the line.

We show the average number of bad splits

Here we explore how often the concept boundaries of the active
concepts in the top layer, \ie where the corresponding pattern begins
and ends, aligns with the location of blank space. Recall that we
remove blank spaces from each line read.

}

\begin{table}[t]
  \centering
    \begin{tabular}{ |c|c|c|c|c|c|c| c|c|c| }     \hline
      & \multicolumn{3}{c|}{ level 1 model }  & \multicolumn{3}{c|}{ level 3 model (Model3) }
      & \multicolumn{3}{c|}{ level 4 model (Model4) } \\  \hline
try, keep & score & bad1 & bad2 & score & bad1 & bad2 & score & bad1 & bad2 \\  \hline
1,1 & 1.64 & 23.5 & 17.2 & 6.9 & 11.3 & 6.7 & 5.6 & 10.6 & 6.6 \\  \hline
2,2 & 1.77 & 23.3 & 17.1 & 8.0 & 10.5 & 6.1 & 7.3 & 9.3 & 5.5 \\  \hline
5,5 & 1.89 & 23.1 & 16.8 & 8.96  & 9.8 & 5.4 & 10.6 & 7.3 & 4.1  \\ \hline
    \end{tabular}
\vspace{.2cm}    
\caption{Average \coma and average count of bad splits (over 188
  episodes) for several levels and beam search parameters. As
  expected, \coma improves with additional search. 
  Number of bad splits and \coma are (negatively) correlated too.}
    \label{tab:splits}
\end{table}

\begin{table}[tbh]
  \centering
  \begin{tabular}{ |c|c|c|c|c| } \hline
    \multicolumn{2}{|c|}{level 2 model} &  \multicolumn{3}{c|}{Model3} \\ \hline
  width 1,1  & width 10,10 & width 1,1 & width 3,3 & width 10,10 \\ \hline
  0.735  & 0.702 & 0.701 &  0.687  & 0.672 \\ \hline
  \end{tabular}
  \caption{ Bad-split ratio scores for a few models and segmentation beam
    widths, average over the same 388 episodes. The bad ratios
    decrease with more training and additional inference
    (width). Multiple runs gives similar results. See also
    Fig. \ref{fig:bad_ratios}. }
\label{tab:bad_ratios}
\end{table}

\begin{figure}[!htbp]
\begin{center}
  \centering
  \hspace*{-0.1cm} \subfloat[Model3: 2 runs with width 1, 2 runs with
    width 10 (\ie 10, 10).  ] {{\includegraphics[height=6cm,width=5cm]
      {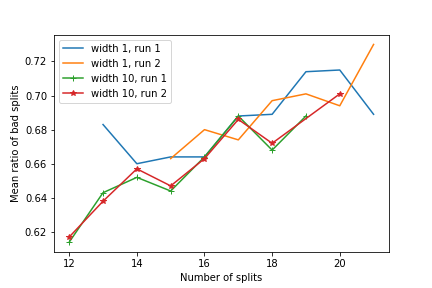} }}
 \hspace*{.5cm}  \subfloat[Model3 and a model tained up to level 2. ]
                    {{\includegraphics[height=6cm,width=5Cm]
                        {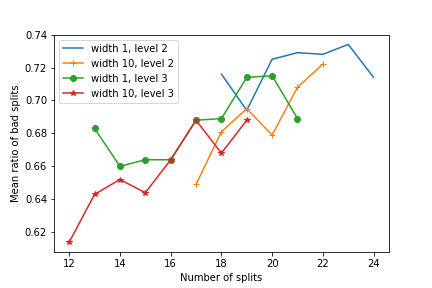} }}
  \hspace*{.5cm}         \subfloat[Model3: Minimum requirement of 5 episodes.   ]
                    {{\includegraphics[height=6cm,width=5cm]
                        {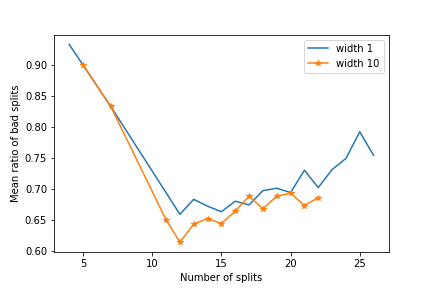} }}
\end{center}
\vspace{.2cm}
\caption{Plots of averages of bad-split ratios for for those split counts
  where we got at least 20 episodes, (a) and (b), or minimum of 5 for
  (c), for the total split count, when a model was run over nearly 400
  episodes. As we increase the beam width or the training, the plots
  move to the left and the bad-split ratios decrease.  }
\label{fig:bad_ratios} %
\end{figure}

That the number of bad splits is decreasing with more training is not
entirely unexpected however, because over time, the newly discovered
concepts correspond to longer strings, and they lead to fewer
overall splits in general (also see Sect. \ref{sec:exps_qloss}). It is
likely that additional inference (with wider width) is more
successful at matching larger concepts, which leads to fewer concepts
in an episode, fewer splits and therefore fewer bad splits (in
addition to improving \comap). We also added the reporting of
average over episodes of the {\bf \em bad-split ratio}, \ie the ratio of bad
splits to total number of splits in an episode, and we have observed
that this ratio also improves somewhat with more training episodes and
layers, as well as with a larger beam width. See Table
\ref{tab:bad_ratios}.  We also averaged the bad-split ratio for each fixed
number of splits separately, and Fig. \ref{fig:bad_ratios} shows the
averages for those splits $k$ for which the number of episodes that
yielded $k$ splits was at least 20 cases (\ref{fig:bad_ratios}(a) and
(b)) or at least 5 cases in \ref{fig:bad_ratios}(c) (to remove clutter
and more easily see the patterns).  For instance, in one run we got 3,
5, 20, and 24 episodes resp. for split counts of 11, 12, 13, and 14,
and among these we are showing averages for 12, 13, and 14 in
\ref{fig:bad_ratios}(c) (average for 11 is not shown as only 3
episodes had that count), and averages for only 13 and 14 are shown
(min count of 20). Multiple runs yield very similar results specially
with higher beam width (a couple shown in \ref{fig:bad_ratios}(a)).
We observe that with higher beam-width and more training the plots
shift to the left (episodes have fewer number of splits), and also the
plots shift down (reflecting a decrease in the bad-split ratio, similar to
Table \ref{tab:bad_ratios}).

A smaller search width for segmentation speeds up inference in each
episode, but the quality of the final selected segmentation can be
poor, and the poor quality may slow down the learning of good
co-occurrences and therefore good compositions in the long run. There
is a tradeoff. In particular, a beam width of say 1 (or a bad
inference algorithm more generally) may introduce too much noise, too
many misteps along the way and by the time a final segmentation is
reached at the highest level (\eg the joining of "nd", of "and",
with "m", from "management", to create "ndm", in level 2 segmentation
by Model4, in Table \ref{tab:seg_examples}).  See for example Appendix
\ref{app:bin} where Expedition is started with 0 and 1 as primitives,
and many (10s of) levels may be required to discover the level of
characters and words. We leave further study of the interaction of the
search width with learning to future work. It will also be informative
to assess the various measures of progress on heldout data.

\subsection{Timings and Computational Costs}
\label{sec:timing}
All code is written in Python.  Each period or sweep of 1500 lines
(episodes) took ~3 minutes, on a Macbook Pro laptop, when layer 1 was
the maximum layer, while it took 30 minutes when the maximum layer was
level 4 (\eg for Model4). We note that one could train models in
parallel and periodically aggregate the models.  Model sizes also grow
with more episodes and layers (additional concepts and edges), from a
few Megs (compressed) when layer 1 is the maximum layer, to low 100s
of Megs for Model4 in our current experiments.  The main time
complexity is in the segmentation search, where concepts try their
holonyms during the search, and predict, for example during \coma
scoring for picking the best segmentation at the top layer. As long as
the data structures are kept relatively small, \eg at most 100s to
1000s of entries (in each of prediction weight maps for each
positions, and part related vertical connections), we expect the cost
of each episode remains manageable. Let $d$ be the maximum over size
of such connection lists (over all concepts). Cost of an episode grows
with product of episode size (number of characters in a line), search
width, number of layers, and (degree) $d$, but we expect that $d$ need
not grow or grow very slowly with the number of concepts. For
instance, see Table \ref{tab:mass} for the case of prediction weights:
we expect, for each concept and position, relatively few edges will
have sizeable weights, and we conjecture that only sizeable weights
(conditional probabilities) are needed for good performance, in
practice. Note that tiny weights also require much more training
(samples) to estimate well.



\co{

  begign buffer predictors, Model3. their top 10...
  
  >code example
  expedition3.Concept.dpw_tabular(Expedition.predictor.begin_buffer_preds[3], poses=[1])
  & "e",0.041 \\ 
 & "a",0.038 \\ 
 & "s",0.033 \\ 
 & "in",0.028 \\ 
 & "the",0.027 \\ 
 & "Thi",0.017 \\ 
 & "and",0.016 \\ 
 & "The",0.015 \\ 
 & "i",0.014 \\ 
 & "t",0.014 \\ 

  level2:
  & "a",0.049 \\ 
 & "i",0.046 \\ 
 & "e",0.042 \\ 
 & "t",0.034 \\ 
 & "s",0.034 \\ 
 & "the",0.029 \\ 
 & "pro",0.027 \\ 
 & "The",0.026 \\ 
 & "T",0.026 \\ 
 & "and",0.020 \\

  level 1:
  & "t",0.070 \\ 
 & "a",0.069 \\ 
 & "i",0.058 \\ 
 & "T",0.053 \\ 
 & "p",0.052 \\ 
 & "e",0.043 \\ 
 & "s",0.034 \\ 
 & "re",0.032 \\ 
 & "co",0.029 \\ 
  & "f",0.028 \\

  & "t",0.085 \\ 
 & "a",0.080 \\ 
 & "s",0.069 \\ 
 & "c",0.067 \\ 
 & "p",0.064 \\ 
 & "i",0.058 \\ 
 & "T",0.054 \\ 
 & "e",0.043 \\ 
 & "o",0.042 \\ 
 & "m",0.040 \\ 

  }

%% file: rel.tex
\section{Related Work}

This work is a continuation of our research on prediction games
\cite{pg1,exp1,complexPatterns08}, sharing the goal of learning a
hierarchy of concepts in a cumulative unsupervised manner. The
motivations and philosophy behind the approach and relations to a few
general learning tasks, \eg distribution or density learning, are
discussed in that work \cite{pg1}. Previously the focus was primarily
on prediction and composition \cite{exp1}, and the need for a more
sophisticated segmentation was later identified but left to future
work \cite{complexPatterns08}.  Segmentation there was a simple left
to right and greedy process: a largest concept (sequence of words or
tokens) was extracted and the resulting segmentation (or tokenization)
was used for self training of prediction weights. To handle more
complex concept structures, it was deemed that a fairly sophisticated
segmentation process would be needed \cite{complexPatterns08}.  A more
elaborate segmentation or inference is also likely needed when noise and
uncertainty is increased, \eg when portions of input are corrupted.
Furthermore, an appropriate smooth objective would also probably be
required to guide the inference.  For this work, therefore, we needed
to both devise a segmentation algorithm and develop an effective
objective to guide the search. The objective had to promote using
larger and possibly recently created (new) concepts, but balanced
against other desiderata, such as the fit with other concepts deemed
present in an episode. The reliance of the segmentation objective on
probabilities and the desire to get better probablities motivated the
use of multiple levels for concepts.

In addition to composition for building higher level concepts
(conjunctions), we have posited that discovery of {\em groupings}
(disjunctions) in concept structure to be important and useful too,
and the two concept creation operations, various forms of disjunctions
and conjunctions, together would create a hierarchy of larger and/or
more abstract concepts \cite{pg1}. This work provides empirical
evidence that creating a composition hierarchy is a practical
possibility. However, learning concepts that also involve grouping or
abstractions of some kinds, remains a challenge and a major open
problem.  We are proposing a kind of structure learning, each concept
a separate structure, thus we seek to learn many structures, without
explicit supervision, and with much sharing of substructure.  The
current concept structures are simple.  Currently, updates are not
performed with respect to a concept's internal structure: updates are
(mostly) limited to simple (scalar) fields and a concept's
associations, \ie the prediction weights.  The system learns immutable
concepts.\footnote{In this implementation, a concept may update its
  parts to contain more than one part-pair, which is a very limited
  structure update.}  In general structure learning is very
challenging (such as learning various subclasses of finite-state
machines, grammar induction, etc.)
\cite{verwer2013PAutomaCAP,Dupont2005LinksBP}. It is an intriguing
question how much learning of additional sophistication in structure
can be effectively supported in the approach presented.

The self-supervised learning in prediction games requires efficiently
predicting and learning prediction weights for a large and a growing set of
concepts in an online manner, and our work on sparse index learning
(or "recall systems") has focused on designing such algorithms and
update techniques
\cite{abound09,updateskdd08,ijcai09,ecir10,extreme2013}.  In
particular, in this work we have focused on simple EMA updates and
simple ways of deriving probabilities from the predictions.

The working of the Expedition system has similarities to n-gram models
for language modeling \cite{slm,manning_nlp}, in that explicit n-grams
are stored for predicting. A major difference is that we aim to learn
a hierarchical vocabulary, while the vocabulary is fixed and flat in
n-gram models, and there is no segmentation task that is integrated
with learning in such systems, to the best of our knowledge. The
prediction in Expedition is also more flexible than common approaches
based on n-grams. Early work on segmenting text include
\cite{Harris1970MorphemeBW}, which used character entropy.

In computer vision, semantic segmentation refers to the task of
assigning classes to pixels, but the classes are often given
\cite{thoma2016,deep_sem_seg21}. In unsupervised (image) segmentation,
pixels are grouped together to form parts of possible objects or
object classes, but no labels are induced.  There is much recent work
in vision in particular that involve various types of self-supervised
learning and proxy or 'pretext' tasks, utilizing various cues such as
visual depth, for feature learning
\cite{Doersch2015UnsupervisedVR,self-depth17,Jing2021SelfSupervisedVF},
\eg using convolution neural networks to build a vocabulary of visual
features.\footnote{We have referred to the representations learned as
  "concepts", \vs "features" (elevated their status), as they are used
  both as predictors as well as targets of prediction, and they
  contain considerable internal structure, such as the parts and
  part-of links, in addition to several scalar fields.  } In the audio
domain, coincidences have recently been used, in part for clustering
to find useful representations \cite{Jansen2020CoincidenceCA}.

Neural networks are universal function approximators
\cite{Hornik1989MultilayerFN}, and with advances of the past decades
(diverse architectures, development and advancement of
backpropagation), and ample data and computation,
have become highly powerful for extracting diverse regularities.
Following the success of neural networks in the vision and speech
domains \cite{alexNet,Hinton2012DeepNN}, large language models via
deep neural networks, using a number of techniques such as embeddings,
prediction, and attention, have had substantial recent successes in
various diverse NLP and related problems
\cite{Collobert2011NaturalLP,Vaswani2017AttentionIA,dong2019unified,
  Rogers2020API, Brown2020LanguageMA}.  In much of existing work in
text, the networks begin with an existing vocabulary and the
embeddings of that vocabulary as input, and it is remarkable that much
powerful learning is achieved without the need to nor the complexity
of segmenting. The meanings of words and patterns can become highly
distributed, providing advantages in making connections among similar
patterns, but also potentially losing some structure and
interpretability. There is also work on character n-gram models based
on neural networks \cite{al2019character}. Sparse mixture of experts
(MOEs), that attempt to activate a small portion of the neural network
on a per example basis (conditional learning or gating), trained via
backpropagation, have had considerable success in further scaling and
speeding up of neural networks training and inference
\cite{fedus2021SwitchTS}.

Concepts are on one hand foundational to human cognition
\cite{bigbook,catDev03,handbookCCS2017}, and are on the other hand
"maddeningly complex"
\cite{bigbook}. Concepts are inter-related in diverse ways
(part-whole, taxonomic, spatio-temporal, domain-specific, ...), or put
another way, concepts seem to enjoy rich "content" (or attributes, in
terms of other concepts).  The nature of concepts and how they are
acquired and adapted over time, along with their rich relations and
flexible use, remains largely a mystery.  There is ongoing debate
among researchers on the various aspects of conceptual development,
for instance, whether concepts develop in a general to specific
manner, or vice versa, the relation between language development and
conceptual change, and whether there are fundamental differences
between say perceptual or more concrete concepts (such as visual
objects, \eg a dog) \vs seemingly more abstract concepts (\eg animal)
\cite{handbookCCS2017,catDev03}.  Researchers have gone as far as
theorizing that children acquire knowledge in a manner very similar to
the progress in science, forming models (theories) that are adapted or
revised with new experience and evidence (the so-called "theory
theory" \cite{theories1}).  Prediction has been proposed as a primary
driver of much of human intelligence \cite{onintelligence}.  Our work
also took inspiration from the neuroidal model of the neocortex and in
particular {\em random access} tasks, \ie tasks that may involve pairs
or multiplicities of concepts (stored items), such as associating two
{\em arbitrary} concepts, from a large space of acquired concepts,
after observing them in one or a few episodes
\cite{circuits}.\footnote{Similar to that work, we have also assumed
  concepts are more "programmable", than the nodes in the common
  neural network models.}  Considering the importance and utility of
concepts for solving advanced information processing tasks, or
symbolic computation under uncertainty, and the complexity of
conceptual phenomena, a diversity of algorithms or (sub)systems,
working together, is likely required \cite{atomsOfNeuralComp}. It is a
major open question whether existing neural network techniques based
on backpropagation, which have now substantially advanced many machine
learning applications, can be extended (\eg perhaps in a posthoc
manner) to support concepts or provide a basis for reaching
human-level cognition.  The extensive research work on the
interpretability of the models learned \cite{carvalho2019ml_interp}
and the related issues of model robustness and brittleness
(adversarial attacks)
\cite{Szegedy2014IntriguingPO,Ilyas2019AdversarialEA}, may also be
linked to the major question of whether neural networks can learn
explicit discernible concepts with some robust internal structure.

%% file: summary.tex
\section{Summary and Future Directions}

We presented and explored a self-supervised learning system,
containing several interacting modules, that acquires a hierarchy of
larger and more specific concepts, from consuming text.  In this
approach, concepts are discrete and connections are sparse, and as we
described, much of the learning (the updates) can be done locally
involving one or two concepts. In particular, concepts do a lot of the
"book keeping" in this approach.  We developed and motivated a
(segmentation) objective, that promotes building and using larger
concepts corresponding to larger patterns in the input.  In
particular, the objective requires (conditional) prediction
probabilities which in turn required some exploration or
experimentation to estimate well.  What the system learns, the
concepts and the prediction weights, are interpretable, and one can
also see what the system "sees" in a given episode (\ie the
segmentation of an input into current concepts).  We explored the
design space and several tradeoffs briefly, such as how to change the
learning rate and whether and when to turn off the learning of new
concepts (compositions), but much work remains in those directions.

An important next extension we hope to develop is in terms of matching
and segmentation (inference) algorithms and objectives to support
approximate matches and segmentations, in order to allow for
additional noise or corruption in the input and to provide more
flexibility in general.  An important open challenge and longer term
area of research is to support some forms of abstraction, internal to
concepts and/or during inference. This connects to the question of how
much further we can push structure-learning in this framework (\eg
learning concept classes that are somewhat closer to the class of
regular expressions or Markov models).  Future directions also include
comparing to existing approaches, and exploring hybrid techniques, as
well as running the system on more diverse data and larger datasets.

%% file: acks.tex
\section*{Acknowledgments}

Thanks to colleagues in the Tetration Analytics group of Cisco, in
particular the paper-reading group members (Vimal Jeyakumar, Weifei
Zeng, Reza Eghbali, Ali Parandehgheibi, ...), for encouragement to
resurrect this work and for discussions and feedback, and to Shashi
Gandham and Navindra Yadav for providing the environment.  Many thanks
to Tom Dean for his encouragement and enthusiasm in the past few
months, and for valuable discussions (the neuroscience side),
suggestions and pointers.

%% file: appendix_rate.tex
\section{Learning Rate Schedules to Fast Approximate Probabilities}
\label{app:dlr}

%








In this section we explore a few properties of EMA and the choice of
learning rate for the prediction edge weights. Several such properties
were also reviewed in \cite{updateskdd08}. The discussion applies to
any relative position with respect to a concept, in particular for the
edges for (relative) position 1, which we assume in our concrete
examples here without loss of generality. We find that starting the
rate $r$ of EMA high and decaying it over time, \eg based on the
number of updates so far, can provide certain advantages in terms of
speed of convergence, and we explain this observation by noting that
EMA reduces to simple averaging (which is the best one can achieve in
a sense), when rate $r$ is decreased in the manner we describe.

An EMA update is a convex combination of the past (summary or
average $a$) and present (latest observation $o$):
\begin{align}a^{(t)} = EMA(a^{(t-1)},
  o^{(t)} , r) = (1-r)a^{(t-1)} + r o^{(t)}, t \in \{1, 2, 3, \cdots\} \end{align} where
$a^{(t)}$ is the (running) average at time $t$ (initially, $a^{(0)}=0$, in case of
prediction weights), $r$ is the learning rate, $0< r <1$, and
$o^{(t)}$ is the observation at time $t$.  For the case of updating
weights for a particular target concept, $o^{(t)}$ is either 0 or 1,
(either that concept is seen in the relative position, or not), \ie a
Bernoulli random variable.  Past observations' weights in the average
reduce exponentially fast as a function of $t$, by $(1-r)^t$,
rendering EMA finite memory in effect, \ie the average of the last
$O(\frac{1}{1-r})$ observations.  The advantage of EMA compared to say
taking a simple average is that only a small constant memory is
required, and with its simplicity, it can track some level of change
or non-stationarity.  Assume after a (predicting) concept $c_1$, a
concept $c_2$ occurs with {\em target probability} of say $0.4$. The
stream of observations that $c_1$ 'sees', with regard to $c_2$ will be
a sequence of 0s and 1s, something like $011000010100\cdots$, \ie
about 40\% of the observations are 1, the rest are 0.  It is not hard
to see that, with a sufficiently low but positive learning rate such
as $r=0.001$, EMA update of edge weights of $c_1$, the weight to
$c_2$, $w_{c_1,c_2,1}$, should converge to the (conditional)
probability $P(c_2|c_1)=0.4$, EMA being a (moving)
average. Fig. \ref{fig:lrs} and Table \ref{tab:dlr} present several
convergence scenarios and properties.  More generally with EMA, the
weights on edges remain non-negative and, moreover, (for a fixed
position) sum to at most 1.0. This property has been referred to as a
sub- or semi-distribution \cite{Dupont2005LinksBP}.  It is also
observed that EMA updating is following a gradient for lowering
quadratic loss in \cite{updateskdd08}.  In general, EMA updates lead
to the edge weights converging to and tracking the approximate
(conditional) probabilities, as long as several conditions are met in
practice, which we describe briefly next.

A first condition or assumption is that there actually exists a target
conditional probability $P(c_2|c_1)$ (\ie it is well defined).  This
condition often holds, \eg in a non-changing fixed corpus of text, and
when episodes are sampled at random. It is not hard to see this is the
case at the primitives or the character level of Expedition in
particular (layer 0). In other cases, $P(c_2|c_1)$ may only slowly
change, and such can be tracked with an appropriate rate.  We note
that there is some non-stationarity in the Expedition system due to
the use of more exploration initially when segmenting (at levels 1 and
up), with a gradual change to more exploitation as concepts are
observed more. We leave a careful understanding of that (challenging)
non-stationarity to future work.  Another condition, alluded to
earlier, is that the learning rate needs to be sufficiently small
compared to the (conditional) probability of interest.  If the
learning rate is set too high, for example if the true target
probability is $P(c_2|c_1)=0.01$ and the learning rate is also set
near or higher than 0.01, there will be too many wide oscillations,
\ie the relative error $\frac{w_{c_1,c_2,1} - P(c_2|c_1)}{P(c_2|c_1)}$
can be greater than say 100\% too often (Fig. \ref{fig:lrs}).
Another practical source of noise or inaccuracy is our size budgets or
limits on number of a concept's connections.  For space efficiency as
well as prediction and update efficiency, we drop edges with small
weights whenever the edge list of a concept becomes too large, \ie
surpasses a size threshold. As long as we are interested in relatively
high weights, \eg near or above $0.01$, a size constraint of say 200
will not cause great inaccuracy (with high probability, \ie with most
co-occurrences). EMA, being a {\em moving} average has the potential
to track some non-stationarity in $P(c_2|c_1)$, but for that, we need
the rate to be reasonably large: if the non-stationarity is too rapid
compared to the size of the rate, EMA is not able to track
it.\footnote{Satisfying multiple goals can be impossible with plain
  EMA: consider two concepts $c_2$ and $c_3$, $P(c_3|c_1)=0.01$ while
  $P(c_2|c_1)$ is substantially larger and moreover oscillates between
  $0.1$ and $0.5$ fairly fast (a few 100s of time points).  Thus,
  learning $P(c_3|c_1)$ well requires a relatively low rate ($r \ll
  0.01$), while learning and tracking the non-stationary $P(c_2|c_1)$
  may require a relatively larger rate, otherwise the estimate may
  converge to near the midpoint.}

\co{
This will introduce further inaccuracies in the conditionals, but with
appropriate size thresholds and learning rates, the error introduced
will be small for relatively high conditional probabilities (edge
weights).

meaning that updates for relative position 1 for a concept $c_1$, lead
to convergence of the edge weights, the edge weight to a concept $c_2$
converging to probability of observing concept $c_2$ immediately after
observing $c_1$, $P(c_2|c_1)$. There several exceptions to
}

\subsection{Rate decay \vs a fixed low rate}

\begin{figure}[!htbp]
\begin{center}
  \centering
  \subfloat[ Level 0, rate experiments.  ]  {{\includegraphics[height=6cm,width=7cm]
                        {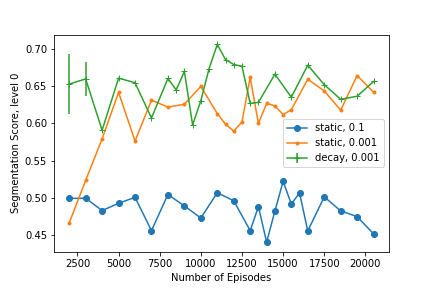} }}
  \subfloat[ Level 1, rate experiments.  ]
           {{\includegraphics[height=6cm,width=7cm]
                        {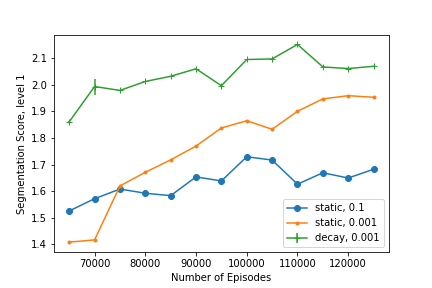} }}
\end{center}
\vspace{.2cm}
\caption{Rate experiments for model training in level 0 and level 1
  (average of 5 and 3 trials respectively). $r_{min}$ of $0.1$ is too
  high and inadequate, even for level 0. Rate decay based on frequency
  has a faster and overall better convergence of \coma compared to a fixed
  rate.  }
\label{fig:rate_coma} %
\end{figure}

When predicting for a target position, we sum the weights on
prediction edges (of all predicting concepts in the context) and
normalize to get a final set of probabilities on the possible
candidate concepts (Figure \ref{fig:scoring}), thus we posit that the
accuracy of the individual edge weights (in the sense described above)
can make a difference in the ultimate prediction performance. For
instance, if the rate is set too high, \eg around $0.01$, we have
observed inferior coherence (\eg see Fig. \ref{fig:rate_coma}).  As
discussed briefly above, setting the rate $r$ of EMA in updating the
prediction edge-weights involves tradeoffs between convergence speed
(speed of learning), accuracy, stability, and adaptability to
non-stationarities.

While a (relatively) low rate is important for learning relatively
small probabilities (the accuracy consideration), the speed of
learning is important too: At any time, the Expedition system may have
many concepts with low frequencies, \eg in the 10s to 100s (seen
counts), which implies their conditional probabilities would be poor
approximations if the learning rates are low. It is important to learn
fast for such large tail of concepts that occur infrequently (but each
episode may contain a few such). Furthermore, ideally we want the
estimates, the prediction weights, to converge fast for those
probabilities that are fairly large (\eg above 0.1), \ie
co-occurrences that are fairly strong. Table \ref{tab:mass} indicates
that many concepts have edges that have those (high) probability
ranges. If we use a low rate, such as 0.001 or 0.0001 (useful for
stability), then it will take a long time to converge for such.  Note
that we have an exploration or optimistic period, set to 50 in the
experiments of this paper, and thus our 'time budget' for learning
probabilities is in the high 10s to low 100s of time steps. We will
see next that starting the rate high and lowering gradually, can lead
to fast convergence specially for the high probabilities fast (\eg
around 0.1), while enjoying the accuracy and stability benefits of a
low rate.

Below, we compare progressively lowering the rate as a function of the
frequency of the updating concept, $c_1$, {\em frequency-based decay},
versus a fixed (low) rate, and we will see that the decay option
speeds up convergence to an error-tolerance region around a target
probability, compared to keeping a fixed (low) rate.  Learning rate
decay has been shown beneficial for training neural nets and there is
research work at explaining the reasons
\cite{You2019HowDL,Smith2018ADA}. Here, we motivate a decay variant in
the context of EMA updates and learning good probabilities fast.

To see why frequency-based decay can perform better than a fixed low
rate, we note that if we lower the rate from 1.0 as a direct function
of the frequency of the (predicting) concept $c_1$, until the minimum
rate $r_{min}$ is reached, we simply get the average of the
observations until $r_{min}$ is reached:

\begin{eqnarray*}
   a^{(t)} &=& EMA(a^{(t-1)}, o^{(t)}, 1/t) = (1-\frac{1}{t}) a^{(t-1)} + \frac{1}{t} o^{(t)} = 
   (\frac{t-1}{t}) \frac{\sum_{i < t}o^{(i)}}{t-1} +\frac{o^{(t)}}{t} \\
   \Rightarrow a^{(t)} &=& \frac{\sum_{i \le t}o^{(i)}}{t},
   \mbox{ \ for \  } 1 \le t \le \frac{1}{r_{min}}    \mbox{\ \ \ \  (EMA is plain average under count-based decay of rate) },
\end{eqnarray*}
where we replaced $a^{(t-1)}$ by $\frac{\sum_{i < t}o^{(i)}}{t-1}$
using induction (holds for $t=1$).  Once $r$ reaches and is fixed at
$r_{min}$, EMA updating becomes effectively averaging with a finite
memory. Simple averaging is the best we can achieve for estimating
probabilities without any extra assumptions, and the frequency-based
decay achieves it. For relatively high probabilities, \eg $p \ge 0.1$,
this can be better than a fixed low rate. Let (relative or
multiplicative) error can be defined as:
\begin{align}
  \label{eq:rele}
  \mbox{error } = \frac{|a^{(t)} - p|}{p} \mbox{ \ (relative error with respect to
    target probability p)},
\end{align}
and we seek this error to be within a tolerance $\eps$, say
$\eps=0.1$.  Note that in general, we require $\theta(\frac{1}{p})$
samples to estimate the probability $p$ well with high
confidence\footnote{Using one form of multiplicative Chernoff bound
  \cite{chernoff52}, with $X$ being sum of $n$ independent random
  variables each in $\{0,1\}$ (Bernoulli), and with $0 < \eps < 1$,
  the probability of large deviation, $ P(|X - \mu| \ge \eps \mu ) \le
  2e^{ - \mu \eps^2 / 3 }$.  With $\mu=np$, as long as $n \ge
  \frac{1}{p}\log(\frac{1}{\delta})\frac{3}{2\eps^2}$, with
  probability at least $1-\delta$ our estimation of $p$ (and $\mu$)
  will be within tolerance $\eps$. A similar lower bound on
  probability of deviation can be derived as well.}  via averaging, (under a
multiplicative error tolerance), as the expected number of
observations to see the first 1 is $\frac{1}{p}$, thus lower
probabilities require longer time to estimate well.  And rate decay
achieves this whether $p$ is large or small (as long $p$ is larger
than $r_{min}$), but EMA with a low fixed rate unnecessarily delays
convergence for the larger $p$.

%
%

\co{
  Complementing the Chernoff upper bound ...
  from math stack exchange, lower bound on the deviation:

 let $X_1 \cdots X_n$ be iid copies of a Bernoulli random variable
 that equals 1 with probability p, thus $\mu_i = p$, $\sigma_i=p(1-p)$,
 $\mu=np$ and $\sigma^2=np(1-p)$

 Using Stirling's formula, show that

%
 \[
 P(|S_n - \mu | \ge \lambda  \sigma) \ge c e^{-C \lambda^2 }
 \mbox{ for some absolute constants $C, c > 0$ and all $\lambda \le c \sigma.$ }
 \]
 
}



\begin{table}[htbp]\center
\begin{tabular}{|c|c|c|c|c|c|c|}     \hline
  & $\pm$errors & $+$errors & 1st time &
  $\pm$errors   & $+$errors & 1st time \\ \hline
  & \multicolumn{3}{|c|}{decay, $r_{min}=0.001$, tolerance $\eps=0.5$} &
  \multicolumn{3}{|c|}{decay, $r_{min}=0.001$, tolerance $\eps=0.1$}  \\\hline
  p=0.25 & 11 $\pm11$ & 4 $\pm9$ & 5 $\pm7$ &
   387 $\pm336$ & 174 $\pm257$  & 32 $\pm81$ \\\hline
   p=0.10 & 39 $\pm39$ & 17 $\pm35$  & 15 $\pm19$ &
   1765 $\pm1022$ & 876 $\pm874$  & 86 $\pm214$ \\\hline
   p=0.05 & 73 $\pm72$ & 34 $\pm65$  & 33 $\pm34$ &
   3492 $\pm1186$ & 1673 $\pm1221$  & 158 $\pm359$ \\\hline
   p=0.01 & 579 $\pm481$ & 341 $\pm437$  & 156 $\pm187$ &
   6820 $\pm932$ & 3039 $\pm1559$  & 566 $\pm862$ \\\hline
    & \multicolumn{3}{|c|}{static $r=r_{min}=0.001$, $\eps=0.5$} &
\multicolumn{3}{|c|}{static, $r=r_{min}=0.001$, $\eps=0.1$}  \\\hline
p=0.25 & 690 $\pm64$ & 0 $\pm0$  & 687 $\pm63$ &
      2468 $\pm483$ & 26 $\pm130$    & 2255 $\pm358$ \\\hline
      p=0.10 & 706 $\pm117$ & 0 $\pm0$ & 697 $\pm116$ &
      3309 $\pm911$ & 363 $\pm491$  & 2135 $\pm532$ \\\hline
      p=0.05 & 701 $\pm161$ & 0 $\pm0$ & 683 $\pm160$ &
      4506 $\pm1164$ & 1126 $\pm957$  & 2008 $\pm616$ \\\hline
      p=0.01 & 1036 $\pm507$ & 157 $\pm315$ & 700 $\pm347$ &
      7083 $\pm858$ & 2293 $\pm1335$     & 1797 $\pm853$ \\\hline
  &  \multicolumn{3}{|c|}{decay, min rate=0.01, $\eps=0.5$} &
  \multicolumn{3}{|c|}{static rate=0.01, $\eps=0.5$}  \\\hline
  p=0.25 & 11 $\pm12$ & 3 $\pm8$ & 5 $\pm8$ &
  73 $\pm21$ & 1 $\pm6$  & 68 $\pm19$ \\\hline    
  p=0.10 & 218 $\pm108$ & 154 $\pm97$ & 17 $\pm25$ &
  258 $\pm107$ & 131 $\pm97$  & 69 $\pm34$ \\\hline
  p=0.05 & 1045 $\pm257$ & 659 $\pm215$ & 37 $\pm41$ &
  1091 $\pm249$ & 601 $\pm225$  & 73 $\pm51$ \\\hline
  p=0.01 & 5050 $\pm366$ & 2229 $\pm459$ & 128 $\pm91$ & 
  5019 $\pm376$ & 2177 $\pm501$    & 95 $\pm100$ \\\hline
\end{tabular}
\vspace{.2cm}
\caption{Counts of when relative errors (Eq. \ref{eq:rele}) are
  too large under two tolerance settings are shown, both
  two-sided ($\pm$errors) and one-sided violations ($+$errors, or overshots), for
  frequency-based decay \vs static learning rate settings, for several
  representative target probabilities,
  averaged over 200 trials. Each trial contains a stream of
  $T=10000$ iid samples of Bernoulli (Boolean, 0 or 1) random
  variables (first row is for $p=0.25$ for drawing a 1, 2nd is with
  p=0.1, etc.).  Learning rate decay to an appropriate minimum rate
  $r_{min}$ converges much faster than a fixed rate at $r=r_{min}$,
  and its errors are symmetric (both over- and under-estimations)
  while, as expected, the static rate consistently under estimates
  until convergence (\ie roughly, when it reaches within tolerance for
  the first time).  }
\label{tab:dlr}
\end{table}

Table \ref{tab:dlr} shows the results of empirical experiments on the
count of violations of tolerance, $\pm$error (averages over 200
trials), \ie when $error > \eps$, for a couple of settings of $\eps$
and the two approaches.  We note that even for a target probability of
$0.01$ (which is closer to $r_{min}=0.001$ than to 1.0 in terms of
ratio), the number of violation of the decay approach is lower.  We
are also reporting the first time point when the estimate falls within
the tolerance region (the first time point that there is no
violation), as a way of assessing the speed of
learning/convergence. Note that this first time $t_0$ is the same as
number of errors until that point when the tolerance region is reached
(before that point, every time point leads to a violation by
definition). After this time $t_0$, there may still be many
violations, specially if the rate is high relative to the target.

The table also gives the number of over-estimation errors (when
$\frac{a^{(t)} - p}{p} > \eps$), and, as expected, the error of
the fixed rate approach are mostly under-estimates, while the errors
of the decay approach are balanced.  We also observe that the count of
errors go up substantially when we raise $r_{min}$, and the errors
continue at a high rate for a target of say $0.01$ after the first
time point. However, the speed of learning does improve, specially for
higher target probabilities, for the fixed rate approach.

\begin{figure}[!htbp]
\begin{center}
  \centering
  \subfloat[Rate $r_{min}=0.001$ ]
           {{\includegraphics[height=7cm,width=4cm]
               {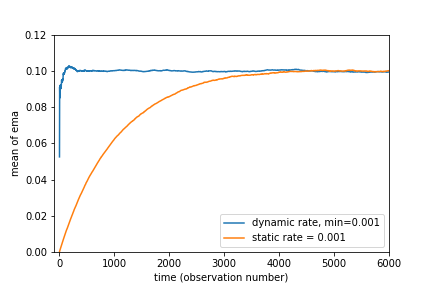} }}
           \subfloat[ Static with percentiles.  ]
                    {{\includegraphics[height=7cm,width=4cm]
                        {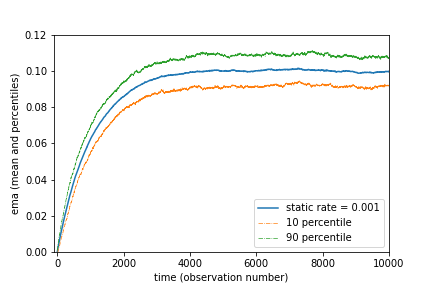} }}
           \subfloat[  Dynamic with percentiles.  ]
                    {{\includegraphics[height=7cm,width=4Cm]
                        {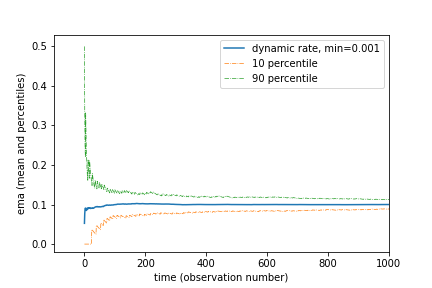} }}
           \subfloat[  Rate $r_{min}=0.05$.  ]
                    {{\includegraphics[height=7cm,width=4Cm]
          {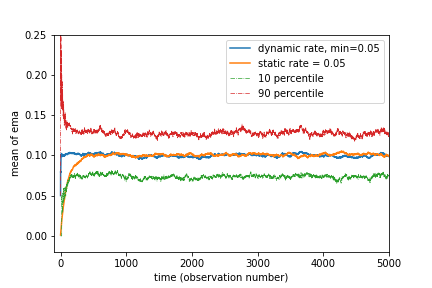} }}
\end{center}
\vspace{.2cm}
\caption{Convergences for EMA under static (fixed at $r_{min}$) and
  frequency-based decay ("dynamic" in the legend) where target
  probability is 0.1. Convergence is faster for the decay
  approach. Increasing $r_{min}$ to 0.05 speeds up the convergence of
  the fixed-rate approach, plot (d), but at the cost of additional
  error. }
\label{fig:lrs} %
\end{figure}

Fig. \ref{fig:lrs} shows similar results in graph format: the
convergence for target probability of $0.1$, additionally showing the 10 and
90 percentile curves.  We note that at any point in time, some 20\%
fraction of probability estimates across all concepts' connections
will be at the 10 or 90 percentile, thus the $r_{min}$ should be low
enough that most probability estimates would not be too far-off from
their target probabilities.

Table \ref{tab:rate_coma} and Figure \ref{fig:rate_coma} show the
segmentation \coma scores at levels 0 and 1 under different rate
schedules. We can see that static at $r_{min}=0.1$ converges fast, but
has inferior performance, even at level 0.  At higher levels, with
many more concepts, smaller probabilities are needed in general and it would
perform even worse.  We observe segmentation scores converge faster
for decay compared to static for $r_{min}=0.001$. After a few thousand
episodes, static eventually converges to a similar performance. We
note that lowering $r_{min}$ to $0.0001$ did not improve performance
at level 0, segmentation performance remaining at an average of $0.65$
(not shown), similar to $r_{min}=0.001$.

We have argued that setting the EMA learning rate as a function of the
frequency of the predicting concept, in particular starting high for
new predictors, can be beneficial. A possible extension is to let the
rate be a function of the newness of the target (predicted) concept as
well, \ie the rate could be somewhat higher for relatively new
concepts.\footnote{A concept that is observed for the first time may
  signify non-stationarity, a new concept that a system will see more
  of, or the concept may just be infrequent. } Depending on the
schedule of when concepts get created and used, this may be a
beneficial extension worth further study.



\co{
\begin{figure}[!htbp]
\begin{center}
  \subfloat[ Moving average of segmentation scores (actual, \ie
    non-optimistic) at level 0, average over 5 runs, shown at a few
    time points, with standard deviations (over the 5 runs).  For
    level 1 experiments, the same model trained at level 0 for 60k
    episodes was used, and performance (segmentation scores) at 65k,
    70k, and 125k are reported. ] {
             \begin{tabular}[b]{|c|c|c|c|}     \hline
         & \multicolumn{3}{c|}{Average performance at a few episodes} \\ \hline
     level 0    & 2000 & 3000  & 5000 \\ \hline
     static, 0.1 & 0.491$\pm$0.044  & 0.499$\pm$0.040  & 0.493$\pm$0.047\\ \hline
     static, 0.001 & 0.465$\pm$0.05 & 0.523$\pm$0.012 & 0.642$\pm$0.058\\ \hline
     decay, 0.001 & 0.659 $\pm$0.043 & 0.660 $\pm$0.023 & 0.661$\pm$0.035  \\ \hline
     level 1 & 65000 & 70000 & 125k \\ \hline
     static, 0.1 & 1.44 & 1.58 &   \\ \hline
     static, 0.001 & 1.42  & 1.43 & 1.95  \\ \hline
     decay, 0.001 & 1.92 &  1.96 & 2.0 \\ \hline
                 \end{tabular}
           }
           \hspace*{0.5cm}
           \subfloat[Segmentation scores (non-optimistic)
             at level 0 and 1 under different rate schedules. ]
           {{\includegraphics[height=6cm,width=6cm]
               {dir_figs/convergence_lev0.png} }}
\end{center}
\vspace{.2cm}
\caption{}
\label{fig:convlev0} %
\end{figure}
}

\begin{table}[htbp]
\begin{center}
\centering
             \begin{tabular}[b]{|c|c|c|c|}     \hline
         & \multicolumn{3}{c|}{Average segmentation performance at a few episodes} \\ \hline
     level 0    & 2000 & 3000  & 5000 \\ \hline
     static, 0.1 & 0.50$\pm$0.03  & 0.50$\pm$0.04  & 0.493$\pm$0.07\\ \hline
     static, 0.001 & 0.47$\pm 0.05$ & 0.52$\pm 0.01$ & 0.64$\pm 0.06$\\ \hline
     decay, 0.001 & 0.66 $\pm 0.04$ & 0.66 $\pm 0.02$ & 0.66 $\pm 0.04$  \\ \hline \hline
     level 1 & 65000 & 70000 & 125000 \\ \hline
     static, 0.1 & 1.53$\pm0.003$ & 1.57$\pm0.03$ & 1.68$\pm0.01$  \\ \hline
     static, 0.001 & 1.41$\pm0.03$  & 1.42$\pm0.02$ & 1.95$\pm0.02$  \\ \hline
     decay, 0.001 & 1.86$\pm0.002$ &  1.99$\pm0.03$ & 2.07$\pm0.06$ \\ \hline
             \end{tabular}
\end{center}
\caption{Moving average of actual (non-optimistic) \coma segmentation scores at
  levels 0 and 1, average over 5 runs, shown at a few time points,
  with standard deviations (over the 5 runs).  For level 1
  experiments, the same model trained at level 0 for 60k episodes was
  used, and performance (\coma scores) at 65k, 70k, and 125k
  are reported. }
\label{tab:rate_coma} %
\end{table}

%% file: appendix_binary.tex
\section{Binary Primitives}
\label{app:bin}

Any learning algorithm and system rests on certain assumptions
regarding the input and has limitations and biases.  The Expedition
system of this paper works primarily bottom up, and if at the lowest
levels the local conditional probabilities are sufficiently close to
random, while there may be rich patterns at sufficiently higher
levels, the system may get stuck, for instance it may stop discovering
new concepts or may build relatively inferior concepts, and it may
never discover the higher regularities.  In this section, we replace
each character with an 8 bit binary code as described
below,\footnote{On this NSF corpus, we have seen 98 unique characters,
  and 7 bits would suffice too.} thus the primitives will be two, 0
and 1. While the text input stream is as before and rich in
regularities, the regularities could be at such a high level that the
system may not discover them via its search of patterns bottom-up
starting from the primitives.  The embedded patterns may go above the
system's head, so to speak.

There are several options for conversion to binary. We chose a
simplest approach: every time a new character is seen  for first
time, we assign it the 8-bit binary code of a (next-available) counter,
initialized to 0, and increment the counter.  Table \ref{tab:bmap}
shows the 8-bit codes for some of the characters (shown in order
increasing counter, or equivalently, roughly in order of when the
system first saw the characters).

\begin{table}
  \begin{tabular}[t]{|c|c|c|c|c|c|}  \hline
    'g'='00000000' & 'o' = '00000001' & 'a'='00000010' & 'l' = '00000010' &  't'='00000100' & 'd' = '00000101' \\ \hline
    'e'='00000110' & 's' = '00000111' & 'c'='00001000' & $\cdots$ & 'I' = '00100111' & '(' = '00101001'  \\ \hline
  \end{tabular}
  \vspace*{1cm}
  \caption{The binary encoding of a few characters, so the word "goal"
    becomes the string "00000000000000010000001000000010" input to the
    system.}
  \label{tab:bmap}
\end{table}

Each episode is a single token (one word) for these experiments,
picked randomly from the random line read (by first splitting the line
by space) so that we can run experiments fast. Each episode is on
average just over 6 characters long, or around 49 bits or binary
primitives long.  The parameters of the system are mostly as rest of
paper: window of 3 concepts both sides for prediction, same
composition parameters, learning rate for prediction being
$r_{min}=0.0001$ and the learning rate schedule is freq. based decay.
We used a larger width of 5 and 15 (5 try and keep 15) for the
segmentation search, as we expected we required more search and
inference to find a good segmentation.  A main difference is guiding
the segmentation: during segmentation we use the {\em segmentation
  credit} (a moving average) for lower-level concepts, to generate
candidate segmentations, but rerank based on optimistic coherence on
top to select a final segmentation from top layer. The segmentation
credit is the historical optimistic coherence for top concepts, and
for lower ones, they get the moving average of the segmentation credit
of concepts they lead to at top layer.  We manually added a layer once
the lower layer's optimistic and non-optimistic coherence converged
(to say within $10\%$ of each other, when averaged over the least
several hundred episodes).

\begin{figure}[htb]
  \centering
  \subfloat[Segmentation scores, binary primitives.]  {{\includegraphics[height=6cm,width=5.2cm]
      {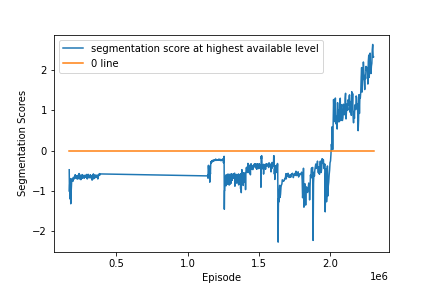} }}
  \hspace*{.1cm}
  \subfloat[Number of concepts per episode in top level.]  {{\includegraphics[height=6cm,width=5.2cm]
      {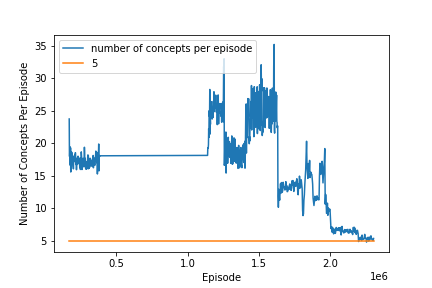} }}
  \hspace*{.1cm}
  \subfloat[Number of concepts with frequency above 50 at episode 2.3mil.]  {{\includegraphics[height=6cm,width=5.2cm]
      {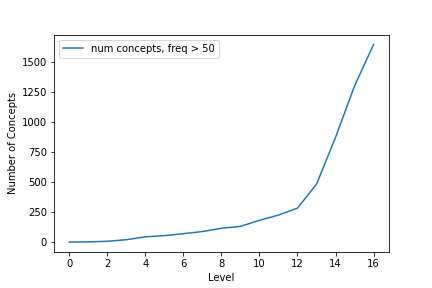} }}
  \vspace{.2cm}
  \caption{(a) \coma scores (actual), when primitives are either 0 or
    1 (when characters are converted to 8-bit strings). The (moving
    average) scores quickly go negative for several initial levels and
    periods, and it appears the system is making no progress overall
    or slow progress. Eventually however (about level 14), as levels
    are added and larger concepts are discovered at those levels, the
    score becomes positive, and progress similar to that on plain text
    appears to start. (b) The number of concepts with freq. above 50,
    at episode 2.3 million, appears to have an inflection point at
    around level 12 or so, and shoots up for higher levels.}
  \label{fig:bprims}
\end{figure}

At the lowest primitives level, initially we get slightly positive
(non-optimistic) segmentation score, around 0.2 (with plain
characters, \coma would surpass 0.6, see for example
Fig. \ref{fig:rate_coma}).  But from then on, as layers 1 and higher
are added in subsequent periods, the average \coma segmentation score
becomes negative all the way till around level 15, as shown in
\ref{fig:bprims}. While the system appears to be stuck, making no
progress in these levels, new concepts continue to get created.
Several of these concepts have positive historical \coma scores
(Eq. \ref{eq:hist}), even though overall, the system gets a negative
score per episode.  Eventually, the score of the system gets back to
the positive region (around level 15).  Fig. \ref{fig:bprims}(b) shows
the number of active concepts at the top level, per episode, and we
note that the number of concepts goes down to about 5, whereas the
average number of character is just over 6 per token input, suggesting
that we are indeed getting bigrams of characters as
well. Fig. \ref{fig:bprims}(c) shows the number of concepts that are
sufficiently seen (above 50 times).  We observe that once level 12 or
13 is added, there appears an inflection point, and the number of
concepts grows rapidly: the system perhaps discovers sufficiently many
concepts corresponding to individual characters, as well as the
bigrams. Before then, the increase in the number of concepts used
appeared slow, and we do not see a large increase in concepts as we
added level, similar to the extent that we saw when we run the system
on plain text.  Note that the ideal minimal system, would discover the
encodings of all characters in level 3 (8 bits, about 100 such for our
corpus), and level 4 would be for the bigrams of such (1000s).

As of this writing, we have continued the training for over 5 million
episodes and through level 19. Actual \coma scores exceed 6 (with try
5, keep 15) and average number of concepts per episode is around 3 in
the top layer. When we give the system the bit string for the
individual characters, \eg one of the 26 lower case letters of
alphabet, for around half of the letters, the system segments them
into a single concept that corresponds to that letter, and segments
the rest into two parts almost always (a beam width of say 5, 5). For
instance, "t", "00000100", is segmented as a single concept, while
"a"="00000010" is often segmented into two parts, "0" and "0000010'.
However, when segmenting its normal input (\ie the binary
representation of individual terms), the splits imposed by the system,
align with the natural splits among the original characters only about
$5\%$ of the time, but additional learning or inference (\eg beam
width) does help reduce the misalignments, the "bad" splits. Still,
from our observations, we don't expect the character-level
misalignments to reduce substantially in this run, and there may be
multiple somewhat equivalent ways of partitioning the stream that lead
to similar \coma performance.  We leave further investigation, in
particular doing multiple runs and experimenting with different beam
widths and other parameters, to future work.


\co{ The coherence goes negative for higher layers!  There are a few
character sequences built (not all possibly 2 bits, but two of them
'00' and '11' at layer 1), and it reaches 10s to low 100s in higher
layers, but coherence so far has remained negative (\eg -0.6, etc)!}

\co{
  
we give the binary rep of a counter, then increment the
counter, for every new char seen. So '00000000' for first char seen (8
bits, but 7 would do too, since we have 94 unique chars), '00000001'
for next, '00000010' for 3rd, etc.

At the primitives level, the co-occurrences may look almost
random. Another challenge is a number of compositions need to be made,
all correctly, before the character level is reached.

It is possible that with its

convert chars to
 binary will it discover the characters and go beyond bigrams of
 characters?? I knew it could be challenging since one assumption has
 been that the sequence should not look random at the low
 level... there should be

--> BTW each episode is one term (i split a line by space, pick a term
at random)..

---> One promising thing is that the number of concepts per episode
goes down with level from 50 concepts (6+ chars), to ~17 at level 6
(lev 5 was 20+, etc).. but i am not sure if the segmentation is good!
(and coherence is still negative at -0.6).. so it may get stuck, or
eventually get stuck.. we'll see...

}